\documentclass[journal]{IEEEtran}
\usepackage{amsmath,amsfonts}
\usepackage{algorithmic}
\usepackage{algorithm}
\usepackage{array}
\usepackage[caption=false,font=normalsize,labelfont=sf,textfont=sf]{subfig}
\usepackage{textcomp}
\usepackage{stfloats}
\usepackage{multirow}
\usepackage{url}
\usepackage{verbatim}
\usepackage{pifont}
\usepackage{graphicx}
\usepackage{cite}
\usepackage{booktabs}
\usepackage{siunitx}
\usepackage{makecell}
\usepackage{xcolor}
\usepackage{tablefootnote}
\usepackage[colorlinks=true,
            linkcolor=blue,
            citecolor=red,
            urlcolor=magenta]{hyperref}
\usepackage{orcidlink}
\newcommand{\xmark}{\ding{55}}

\usepackage[utf8]{inputenc}
\usepackage{textgreek}
\DeclareUnicodeCharacter{03BC}{\textmu}

\begin{document}

\title{AgenticSizing: A Large Language Model-based \\ Multi-Agent Framework for Analog Circuit Sizing}

\author{
    Yijia Hao \textsuperscript{\orcidlink{0000-0002-0491-8201}}, 
    Pratibha Verma \textsuperscript{\orcidlink{0000-0001-9237-1789}}, \IEEEmembership{Member,~IEEE}, 
    Dongxu Guo \textsuperscript{\orcidlink{0009-0001-8446-531X}}, \\
    Cristian Sestito \textsuperscript{\orcidlink{0000-0002-7731-0002}}, \IEEEmembership{Member,~IEEE}, 
    Michael O'Boyle \textsuperscript{\orcidlink{0000-0003-1619-5052}}, \IEEEmembership{Fellow,~IEEE}, \\
    Christos-Savvas Bouganis \textsuperscript{\orcidlink{0000-0002-4906-4510}}, \IEEEmembership{Senior Member,~IEEE}, 
    and Themis Prodromakis \textsuperscript{\orcidlink{0000-0002-6267-6909}}, \IEEEmembership{Senior Member,~IEEE}

\thanks{This work was supported by the Engineering and Physical Sciences Research Council (EPSRC) AI Hub for Productive Research and Innovation in eLectronics (APRIL) under Grant No. EP/Y029763/1, and by the Royal Academy of Engineering (RAEng) Chair in Emerging Technologies under Grant No. CiET1819/2/93.
\textit{(Corresponding author: Cristian Sestito)}.

Y. Hao, D. Guo, C. Sestito and T. Prodromakis are with the Centre for Electronics Frontiers, Institute for Integrated Micro and Nano Systems, School of Engineering, The University of Edinburgh, UK (email: csestito@ed.ac.uk).

P. Verma contribued to this work while affiliated with the Centre for Electronics Frontiers, Institute for Integrated Micro and Nano Systems, School of Engineering, The University of Edinburgh, UK. Current affiliation: Department of Electrical Engineering, Indian Institute of Technology Indore, India.

M. O'Boyle is with the School of Informatics, The University of Edinburgh, UK.

C.S. Bouganis is with the Department of Electrical and Electronic Engineering, Imperial College London, UK.}
}

\maketitle

\begin{abstract}
Analog circuit sizing remains a challenging and time-consuming task due to the large design space, strong performance trade-offs, and increasing circuit complexity in scaled technologies. Although recent large language model (LLM)-based methods show promise in improving sample efficiency and interpretability, existing approaches often lack explicit circuit-topology understanding and are mainly evaluated on relatively simple analog building blocks. This paper presents a multi-agent LLM-based framework for complex analog circuit sizing. The proposed framework first analyzes the circuit topology and decomposes the netlist into functional blocks and substructures. It also extracts lightweight design knowledge for reuse. Based on the extracted topology and knowledge, a planner coordinates multiple role-specialized sizing agents to update design variables and achieve global performance specifications. This workflow mimics the collaborative process of an expert analog design team and provides a structured, interpretable, and simulation-driven optimization procedure. The framework was validated on eight circuits, with the largest design containing up to 55 transistors and 60 sizing variables. Notably, for the LDO benchmark, the proposed method achieved a 60\% success rate with an average of 83 iterations, where classical optimizers failed to find feasible solutions. Further, ablation studies demonstrate that topology understanding, design-knowledge infusion, and agent specialization provide complementary benefits. The source code is available to support reproducibility.
\end{abstract}


\section{Introduction}
\IEEEPARstart{A}{nalog} circuits work as an essential interface between digital processing systems and the physical world. Despite their critical importance, the transistor sizing process remains time-consuming due to the vast design space and the inherent trade-offs among performance metrics, such as gain, bandwidth, power consumption, noise, and linearity. Consequently, the design process still heavily relies on human intuition and experience. Furthermore, as technology nodes continue to scale down, the complexity of analog circuit design increases correspondingly, posing significant challenges for designers to simultaneously meet aggressive time-to-market schedules and stringent performance specifications.

To improve the efficiency of the analog design flow, electronic design automation (EDA) techniques have evolved alongside conventional manual design methodologies \cite{899053,102664,11187385}. Early approaches were primarily optimization-based methods, such as evolutionary algorithms \cite{6466380,9149787,LIU2009137,10.1145/1529255.1529264}, which incorporate AI into the transistor-sizing process. More recently, surrogate model-based approaches have been explored, in which machine learning models are trained online to predict circuit performance and guide the search process. For example, Bayesian optimization (BO) with Gaussian process regression (GPR) has been widely investigated \cite{8714788,6714475,8116661}, as it can reduce the number of expensive circuit simulations required by evolutionary algorithms. However, its efficiency degrades as the design space expands and the number of performance trade-offs increases. Reinforcement learning (RL) \cite{10614385,10443951,11063353} has also been explored as an alternative framework, in which neural networks learn sizing policies by interacting with the simulator and receiving rewards based on performance metrics, with the goal of maximizing the expected cumulative reward. Although such policies may be transferable across related design tasks, RL methods generally suffer from low sample efficiency \cite{9586139}. Overall, both surrogate model-based optimization and RL-based frameworks require a large number of simulations to obtain high-quality designs, particularly for complex analog circuits. In addition, their limited transparency remains a significant challenge, as the underlying models are often treated as black boxes, making it difficult for designers to assess the reliability of the obtained solutions.

More recently, large language model (LLM)-based approaches \cite{yin_ado-llm_2024,liu_eesizer_2026} have attracted wide interest due to their few-shot learning, in-context reasoning capabilities and the prior knowledge embedded in pre-trained cloud-based models. These methods exhibit advantages in terms of sample efficiency and interpretability over alternative frameworks. However, LLMs still face two challenges that hinder their application in analog circuit sizing. First, the circuit understanding of existing LLM-based approaches remains limited. The circuit topology is typically provided in the form of a netlist without annotations, which makes it difficult to determine whether the LLM truly comprehends the underlying sizing problem. Consequently, the optimization process often resorts to random or heuristic exploration. Furthermore, the general knowledge embedded in these models lacks the specific domain expertise required for analog circuit design, thereby constraining their design efficiency. Second, existing LLM-based sizing frameworks have been primarily evaluated on basic analog building blocks. However, practical analog systems can readily exceed 50 transistors \cite{10229975,10472895}. Liu \emph{et al.} reported a 60\% success rate for a medium-complexity amplifier with 20 transistors \cite{11318860}. This result indicates that direct LLM-based sizing may have limited scalability to highly integrated circuits.

To address these challenges, this work introduces AgenticSizing, a multi-agent optimization framework for analog circuit sizing. Inspired by existing analog design workflows, the framework decomposes complex circuits into functional blocks and coordinates role-specific agents to meet global performance objectives. It integrates circuit topology information and reusable design knowledge to guide optimization, improving both sizing efficiency and interpretability. Rather than using the LLM as a standalone optimizer, AgenticSizing investigates how structured agent coordination and circuit-level context can support effective analog sizing across both commercial EDA and open-source environments. We evaluate the proposed framework on eight circuits spanning different levels of complexity and compare its performance with three baseline methods. For the most complex circuit, the proposed framework achieves a 60\% success rate with an average of 89.7 simulation evaluations, whereas all baseline methods fail to find a feasible design within the six-hour time limit. The key contributions of this work are as follows:

\begin{itemize}
\item We propose an LLM-based decomposition tool that leverages circuit topology information to decompose complex analog circuits into functional blocks and substructures. The software is open-source at \href{https://github.com/aprilaihub/agentic-analog-sizing}{https://github.com/aprilaihub/agentic-analog-sizing}.
\item The framework leverages an LLM-based knowledge extraction tool that captures performance trade-offs, functional role–performance relationships, and design parameter–performance trade-offs, enabling the extracted knowledge to be reused across different circuits.
\item We introduce a novel multi-agent LLM workflow that emulates the collaborative process of expert analog design teams by incorporating structured design knowledge and topological understanding.
\item We demonstrate the superior performance of the proposed framework over competitive optimization approaches on medium- to high-complexity benchmarks and quantify the individual impact of each contribution.
\end{itemize}

The remainder of the paper is organized as follows. Section II reviews LLM-related work in analog circuit sizing. Section III illustrates the methodology and components. Section IV presents experimental results from low- to medium-complexity circuits, extending to a high-complexity case, along with an ablation study. Section V concludes the paper.

\section{Related Work in LLM-based Analog Circuit Sizing}

\begin{table*}[!t]
\centering
\setlength{\tabcolsep}{2pt}
\caption{Comparison of representative LLM-enabled analog design frameworks.}
\label{tab:llm_framework_comparison}
\resizebox{\textwidth}{!}{%
\begin{tabular}{l c c c c c c c c c c c}
\toprule
\multirow{3}{*}{Framework}
& \multicolumn{3}{c}{Benchmark}
& \multicolumn{5}{c}{LLM-Enabled Methodology}
& \multicolumn{3}{c}{Deployment} \\
\cmidrule(lr){2-4}
\cmidrule(lr){5-9}
\cmidrule(lr){10-12}

& \makecell{Benchmark\\Breadth}
& \makecell{Largest Reported\\Circuit}
& \makecell{Max.\\Variables}
& \makecell{Closed-Loop\\Simulation}
& \makecell{Structural\\Reasoning}
& \makecell{Knowledge\\Memory}
& \makecell{Multi-Agent\\Coordination}
& \makecell{Adaptive\\Search}
& \makecell{Commercial\\EDA}
& \makecell{Open-\\Source}
& \makecell{Cross-\\Node} \\
\midrule

EEsizer\cite{liu_eesizer_2026}
& 6 circuits & 20 transistors & N/R
& \ding{51} & \ding{55} & \ding{55} & \ding{55} & \ding{55}
& \ding{55} & \ding{51} & \ding{51} \\

AnalogCoder\cite{lai_analogcoder_2025}
& 24 circuits & N/R & N/R
& \ding{51} & \ding{55} & \ding{55} & \ding{55} & \ding{55}
& \ding{55} & \ding{51} & \ding{55} \\

AnaFlow\cite{ahmadzadeh_invited_2025}
& 2 circuits & 16 transistors & N/R
& \ding{51} & \ding{51} & \ding{55} & \ding{51} & \ding{55}
& \ding{55} & \ding{55} & \ding{55} \\

AmpAgent\cite{liu2024ampagent}
& 7 circuits & N/R & N/R
& \ding{51} & \ding{55} & \ding{51} & \ding{51} & \ding{51}
& \ding{51} & \ding{55} & \ding{51} \\

LEDRO\cite{LEDRO}
& 22 circuits & 19 transistors & 20 variables
& \ding{51} & \ding{55} & \ding{55} & \ding{55} & \ding{51}
& \ding{51} & \ding{51} & \ding{51} \\

LLM-USO\cite{somayaji_llm-uso_2026}
& 24 circuits & N/R & N/R
& \ding{51} & \ding{55} & \ding{51} & \ding{51} & \ding{51}
& \ding{55} & \ding{55} & \ding{55} \\

\textbf{This work}
& \textbf{\makecell{8 circuits\\(3 categories)}}
& \textbf{55 transistors}
& \textbf{60 variables}
& \ding{51} & \ding{51} & \ding{51} & \ding{51} & \ding{55}
& \ding{51} & \ding{51} & \ding{55} \\

\bottomrule
\end{tabular}%
}

\vspace{1pt}
\parbox{\textwidth}{\footnotesize
$\checkmark$: explicitly supported;
\xmark: unsupported or not reported;
N/R: quantitative benchmark information not reported.
}
\end{table*}

Analog circuit sizing aims to determine transistor dimensions and component values that satisfy target performance specifications. Several LLM-based multi-agent frameworks have recently been proposed as alternatives or complements to conventional optimization routines. AnaFlow \cite{ahmadzadeh_invited_2025} uses four agents, namely explainer, DC reviewer, specs reviewer, and sizer, to support reasoning-driven and explainable sizing with iterative simulation feedback. Liu et al. \cite{liu_large_2025} adopt a two-agent hierarchy consisting of an expert for high-level reasoning and an employee for data extraction, focusing on extracting sizing relationships from circuit documentation. AmpAgent \cite{liu2024ampagent} uses three sequential agents: a literature analysis agent retrieves formulas and stability conditions, a mathematics reasoning agent derives stage-level design targets, and a device sizing agent uses conventional optimizers to size multi-stage amplifiers. LLM-USO \cite{somayaji_llm-uso_2026} is a hybrid agentic framework in which an LLM agent and a BO algorithm propose design points through a shared simulation buffer. It uses a working LLM generates transferable knowledge summaries, while a critique LLM refines them for reuse across circuits. Circuit-Agent and EEsizer \cite{yang_circuit-agent_2025, liu_eesizer_2026} also use multi-agent architectures, including selector, extractor, calculator, composer, and validator roles, with an emphasis on integrating LLM reasoning with external EDA tools for circuit evaluation and verification.

As summarized in Table \ref{tab:llm_framework_comparison}, existing LLM-enabled analog design frameworks commonly support closed-loop simulation, but differ substantially in reasoning capability, benchmark scale, and deployment support. Methodologically, multi-agent coordination is supported by only a subset of frameworks, and explicit structural reasoning and knowledge bases are rarely incorporated. In terms of benchmark scale, the reported breadth ranges from 2 to 24 circuits, and the largest reported circuit in prior work contains at most 20 transistors.

These studies show that LLMs can provide useful reasoning interfaces for analog sizing and can be coupled with simulation tools to support iterative refinement. The existing literature, however, remains at an early stage. Prior work has primarily explored how to organize LLM agents, prompt them with circuit information, and connect them to external EDA tools. Less attention has been given to circuit representation during optimization and to the retention and transfer of design knowledge across sizing tasks. Furthermore, the applicability of LLMs to complex analog circuits remains largely unexplored. This limitation is significant because practical analog circuits typically comprise multiple interconnected devices and circuit blocks, resulting in a high-dimensional and strongly coupled design space. Effective optimization therefore requires representations that capture circuit topology, device roles, and interconnections, together with mechanisms for retaining and reusing design knowledge to guide the search process and reduce redundant simulations. 

Motivated by these observations, this work investigates agentic LLMs for simulation-in-the-loop analog circuit sizing under complex design settings. Compared with prior frameworks, AgenticSizing evaluates eight circuits spanning three representative analog circuit categories: operational transconductance amplifiers (OTAs), bandgap references (BGRs), and low-dropout regulators (LDOs), with up to 55 transistors and 60 design variables.

\section{Methodology}
In this work, we adopt agentic LLM principles \cite{yao2023reactsynergizingreasoningacting, schick2023toolformerlanguagemodelsteach} for analog circuit sizing by coordinating role-specific agents through structured state transitions and tool interactions. Since no LLM is specifically developed for analog circuit design, and training such a model would require substantial domain-specific data, ChatGPT-5.2 is used as the underlying LLM agent, following prior LLM-based EDA research \cite{yin_ado-llm_2024,liu_eesizer_2026}.

\subsection{The AgenticSizing Workflow} 
\begin{figure*}[!htb]
\centering
\includegraphics[width=7in]{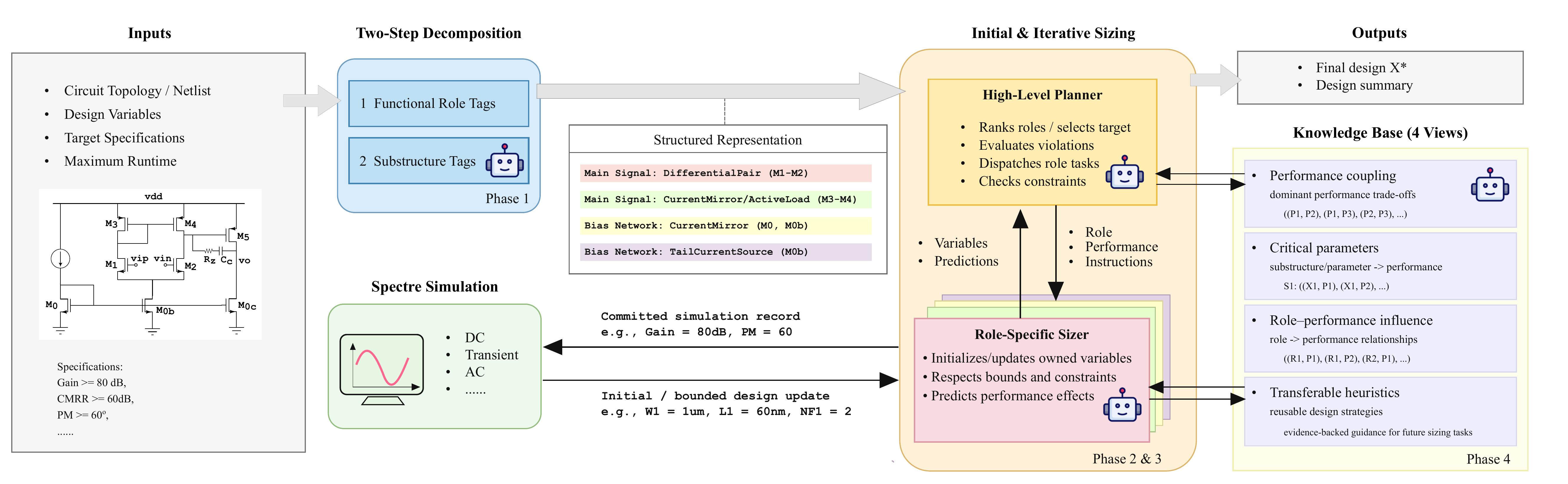}
\caption{Four-phase agenticSizing framework for automated analog IC sizing. The proposed method integrates LLM decomposer (phase 1), LLM planner and sizer (phase 2 and phase 3), and LLM knowledge extractor (phase 4).}
\label{flow}
\end{figure*}

The AgenticSizing workflow, illustrated in Fig.~\ref{flow}, is a multi-agent framework requiring only the circuit netlist, design-variable bounds, target specifications, simulator interface, and runtime budget. Circuit-specific prompt content is generated automatically from topology tagging, simulation history, and knowledge-base retrieval, allowing the same agent templates to be reused without circuit-specific prompt engineering. The workflow partitions the circuit into functional roles, generates an initial design candidate, iteratively refines it through planner--sizer interactions, and extracts reusable knowledge from the simulation results. The following subsections describe these stages in detail.

\subsubsection{Two-step decomposition}
The first stage uses a decomposer agent to convert the transistor-level circuit netlist into a structured representation for use by subsequent agents. Details of the decomposition are presented in Section III-B. The output of this stage defines the ownership of each design variable and provides the structural vocabulary used by the planner, the sizers, and the knowledge base.

\subsubsection{Initial Sizing}
After decomposition, the framework constructs an initial design point. A high-level planner initiates the sizing process by coordinating the role-specific sizing agents according to the design specifications, decomposed circuit structure, and available prior knowledge. The sizers independently assign initial values to their role-scoped design variables while meeting the specified bounds and constraints. These role-level assignments are then assembled into a complete design vector and evaluated by the simulator. The resulting performance vector is recorded as the initial simulation result, providing the starting point for the subsequent iterative optimization process.

\subsubsection{Iterative Sizing}
The iterative sizing stage performs closed-loop refinement of the initial design. At each iteration, the current design is evaluated against the target specifications, and the planner coordinates the selection of a role-specific sizing action to address the remaining performance gaps. The selected sizer then updates only its assigned design variables within the specified bounds and submits the resulting candidate design for simulation. The newly obtained performance vector is appended to the simulation history and used to guide the next iteration. This process alternates between planner-guided design updates and simulator-based evaluation until the design satisfies the target specifications or the optimization process terminates.

\subsubsection{Knowledge extraction}
Knowledge extraction converts empirical sizing experience into reusable symbolic knowledge. After the iterative optimization completes, the LLM knowledge extractor analyzes the relationship between design-variable changes, circuit substructures, functional roles, and performances.
The extracted knowledge is then stored in four complementary views. More details can be found in Section III-D. This knowledge base is then reused by the planner and sizers in subsequent optimization tasks. In this way, the framework accumulates structured design experience that improves future role selection, variable tuning, and trade-off reasoning.

\subsection{LLM decomposer}
\begin{figure}[!t]
\centering
\subfloat[]{%
  \includegraphics[width=0.48\columnwidth]
  {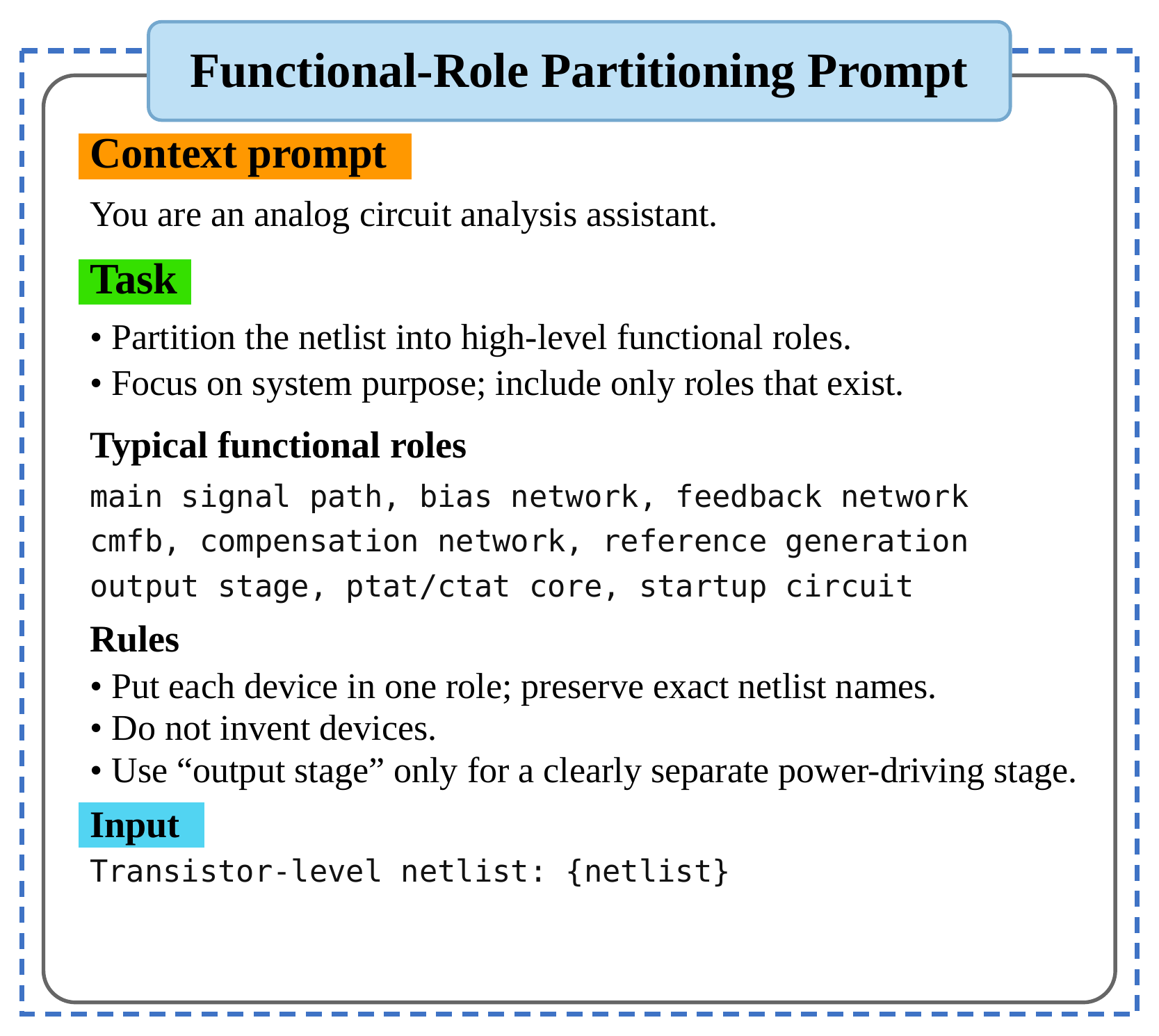}}
\hfill
\subfloat[]{%
  \includegraphics[width=0.52\columnwidth]
  {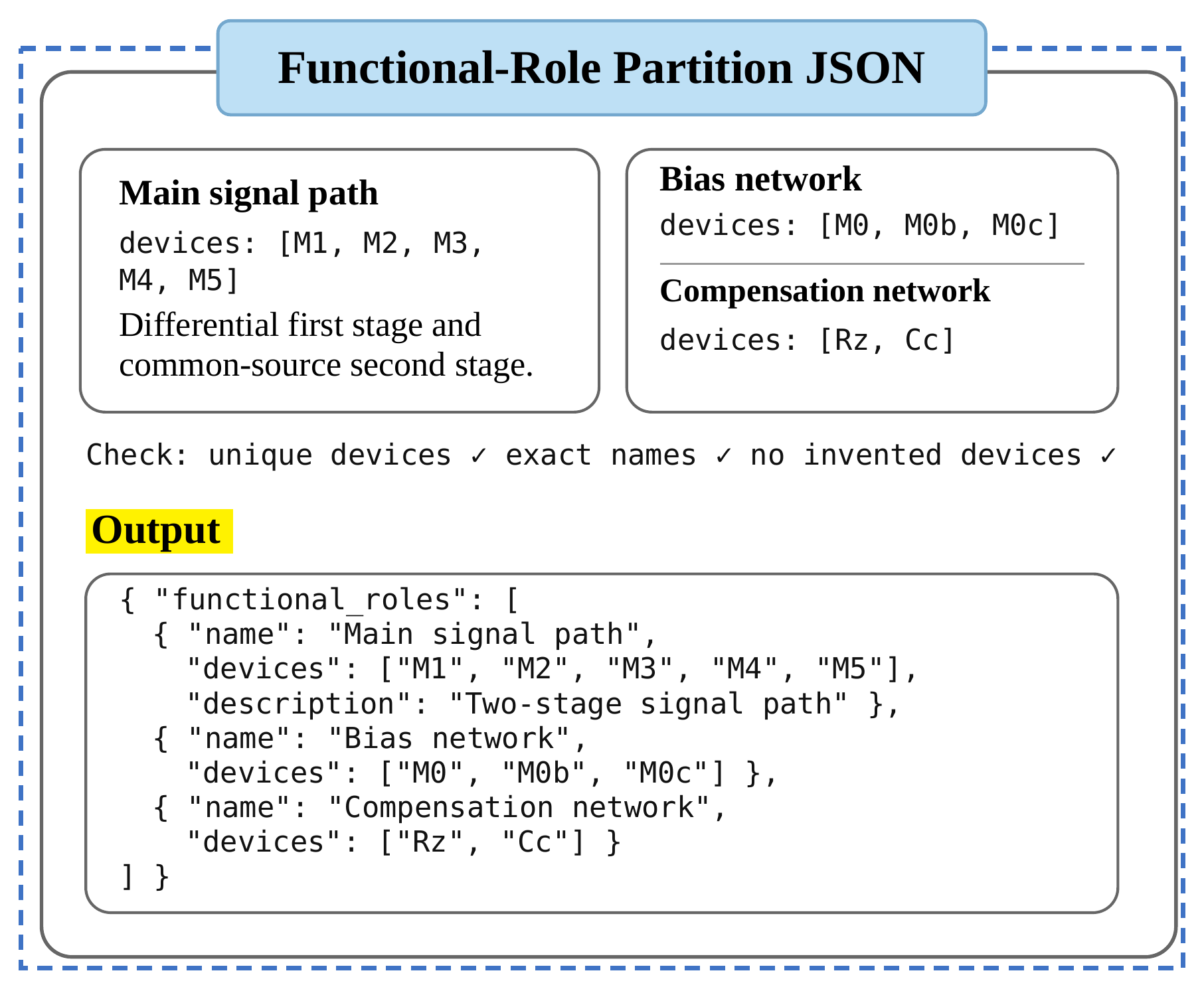}}

\subfloat[]{%
  \includegraphics[width=0.48\columnwidth]
  {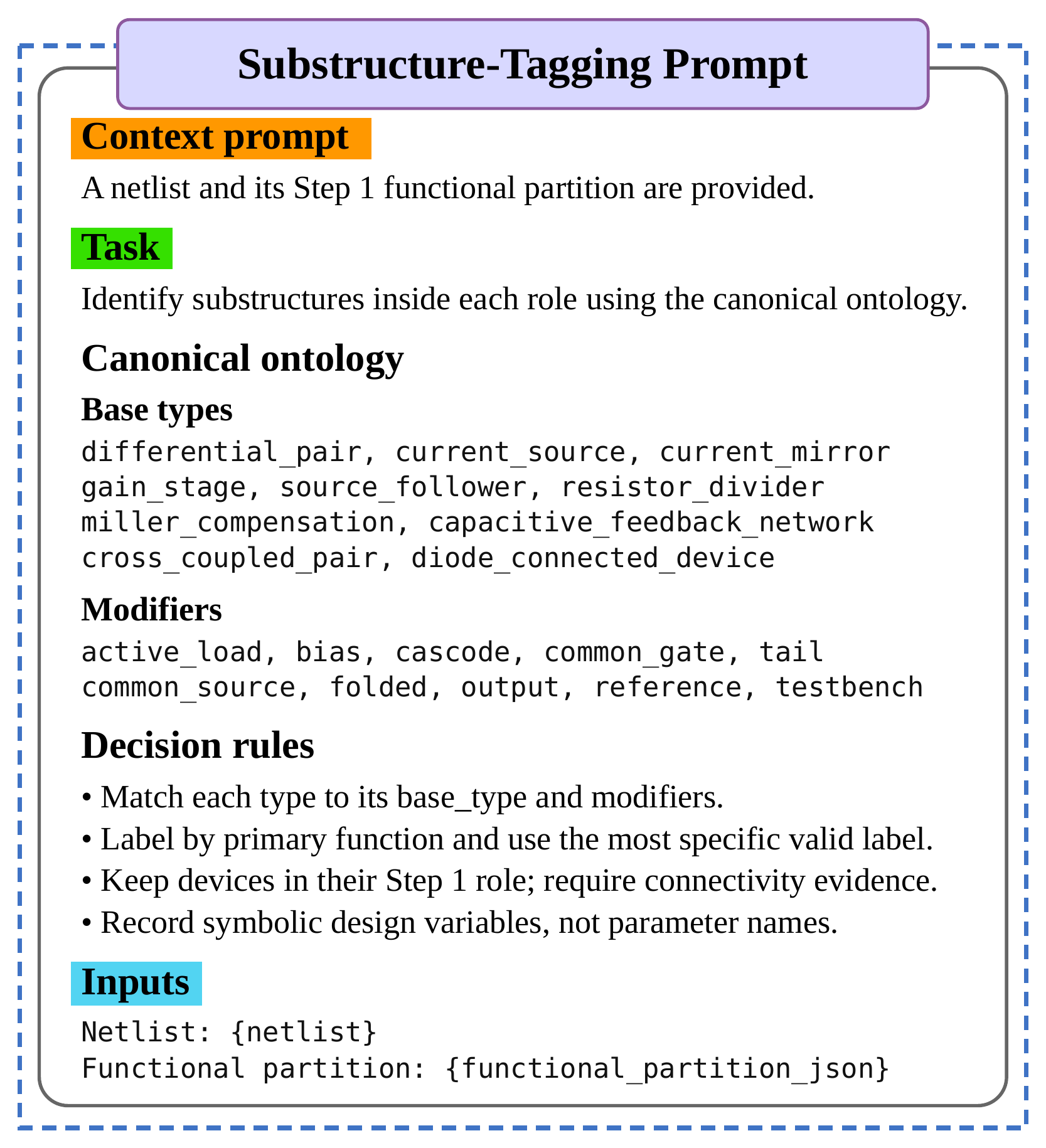}}
\hfill
\subfloat[]{%
  \includegraphics[width=0.52\columnwidth]
  {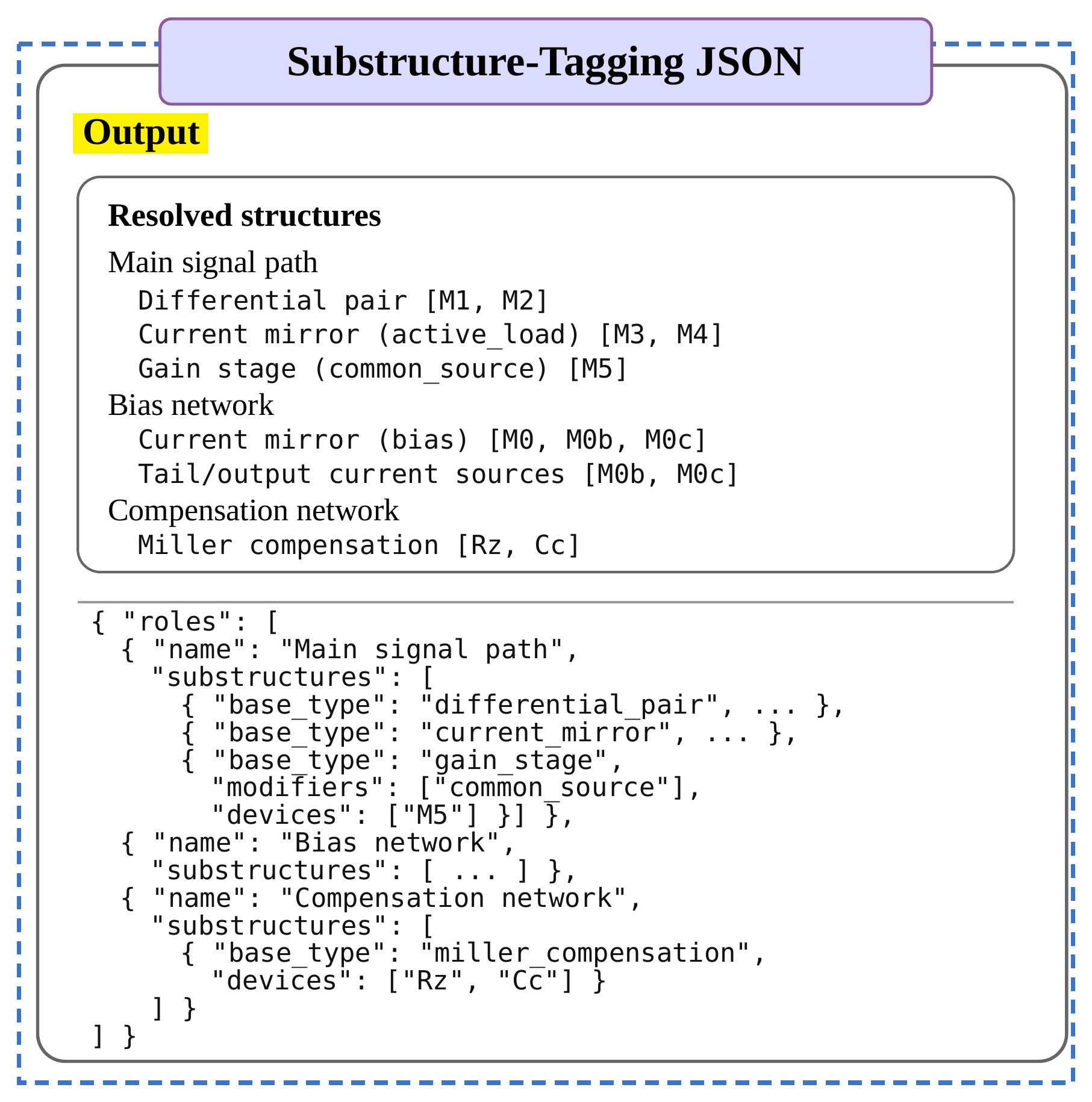}}
\caption{Overview of the two-step circuit decomposition process:
(a) functional-role partitioning prompt;
(b) structured JSON representation of the identified functional roles;
(c) substructure-tagging prompt; and
(d) structured JSON representation of the constituent substructures.}
\label{decomposer}
\end{figure}

The AgenticSizing workflow uses four specialized agents. The LLM decomposer agent translates transistor-level analog netlists into a normalized symbolic representation of circuit function and topology, serving as a bridge between low-level device connectivity and high-level design reasoning. The tool follows a two-stage LLM-assisted pipeline. In the first stage, the decomposer agent partitions the circuit into predefined high-level function roles, such as main gain stage, bias network, feedback network, output stage, startup circuitry, etc.. Each device is strictly assigned to one role. In the second stage, the tool decomposes each functional role into canonical circuit substructures drawn from a controlled ontology. Each substructure is represented by a canonical label, base type, modifiers, connectivity evidence, and associated design variables (Fig.~\ref{decomposer}). The ontology is defined once using common analog primitives, including differential pairs, current mirrors, resistor dividers, and cross-coupled pairs, while modifiers such as \emph{cascode}, \emph{folded}, and \emph{tail} provide further specialization. For example, reference and tail current sources share the base type \emph{current source} but use the \emph{reference} and \emph{tail} modifiers, respectively. Restricting the LLM to these predefined terms normalizes synonymous descriptions and prevents unsupported labels, thereby supporting consistent knowledge retrieval and transfer across circuits. The ontology is extensible and requires no circuit-specific modification for the evaluated benchmarks.

\subsection{LLM Planner and Sizers}
The sizing is performed by LLM planner and sizers. The system follows a $1+N$ planner-sizer pattern, where a single planner agent performs global planning and coordination, while multiple role-scoped sizer agents propose parameter updates for the circuit roles identified by the decomposer tool. 

\begin{figure*}[!t]
\centering
\includegraphics[width=6.5in]{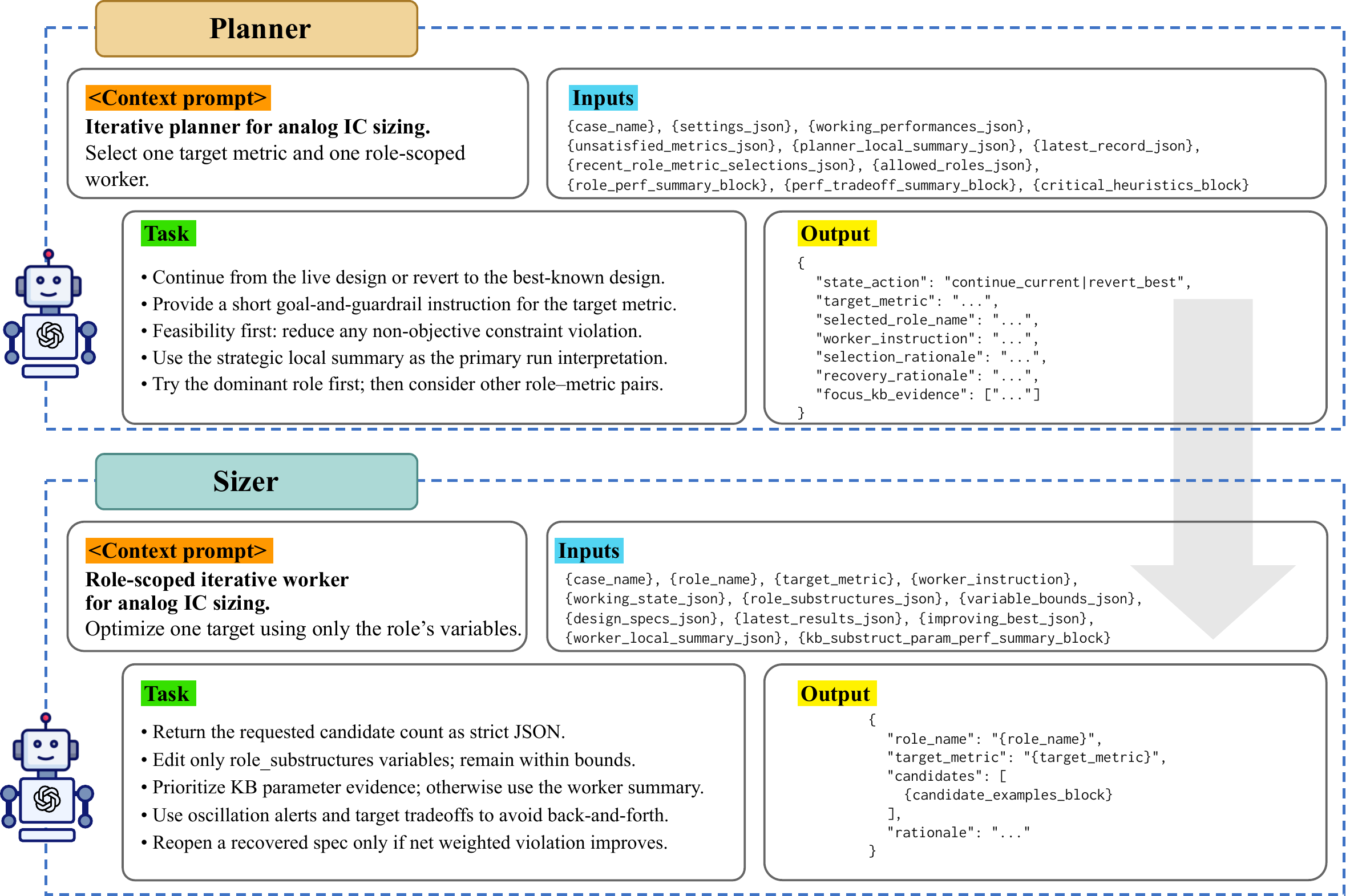}
\caption{Overview of the AgenticSizing planner and sizer. Planner selects metric–role pairs based on the current design state, performance objectives, and historical sizing records. Sizer generates role-specific sizing candidates using bounded variable updates, knowledge-base evidence, and historical worker summaries.}
\label{fig_6}
\end{figure*}

The planner acts as a high-level decision agent. During initialization, it ranks all available functional roles and produces an ordered execution plan, where each role receives a priority, a set of focus metrics, and a worker instruction. During iterative sizing, it evaluates the latest simulation results, identifies unsatisfied specifications, and dispatches one role sizer to optimize one target metric (Table~\ref{tab:prompt_components}). Before prompt construction, the knowledge base is filtered according to the active specifications, such that only relevant role--performance relations, performance tradeoffs, and design heuristics are provided to the planner. The planner also decides whether to continue from the current design or revert to the best-known design when recent evidence suggests an unproductive trajectory. Each sizer performs local design-variable optimization within its assigned role. It receives the planner instruction, role substructures, variable bounds, recent simulation records, improving-design examples, and knowledge filtered for the selected target metric and functional role (Table~\ref{tab:prompt_components}). Then the sizers propose updates only for variables owned by its assigned role. Fig.~\ref{fig_6} illustrates representative prompts and responses of the planner and sizer agents during iterative sizing.

\begin{table}[!t]
\centering
\scriptsize
\caption{Prompt context components for planner and sizer agents.}
\label{tab:prompt_components}
\setlength{\tabcolsep}{2.3pt}
\begin{tabular}{l l l l}
\hline
\textbf{Agent} & \textbf{Component} & \textbf{Source} & \textbf{Purpose} \\
\hline

\multirow{5}{*}{Planner}
& Design specifications & YAML input & Define optimization goals \\
& Tagging results       & Tagging module & Define role decomposition \\
& knowledge base summary            & Knowledge base & Provide heuristics and priors \\
& Recent run history    & Simulation records & Reflect search progress \\
& Local state summary   & Planner memory & Preserve decision context \\
\hline

\multirow{6}{*}{Sizer}
& Role and target metric & Planner output & Define local objective \\
& Role-scoped variables  & Tagging module & Restrict editable variables \\
& Variable bounds        & Config file & Keep updates feasible \\
& Knowledge base evidence            & Knowledge base & Guide parameter updates \\
& Improving examples     & Simulation records & Ground candidate proposals \\
& Local state summary    & Worker memory & Preserve refinement context \\
\hline

\end{tabular}
\end{table}

\begin{figure}[!t]
\centering
\includegraphics[width=3.45in]{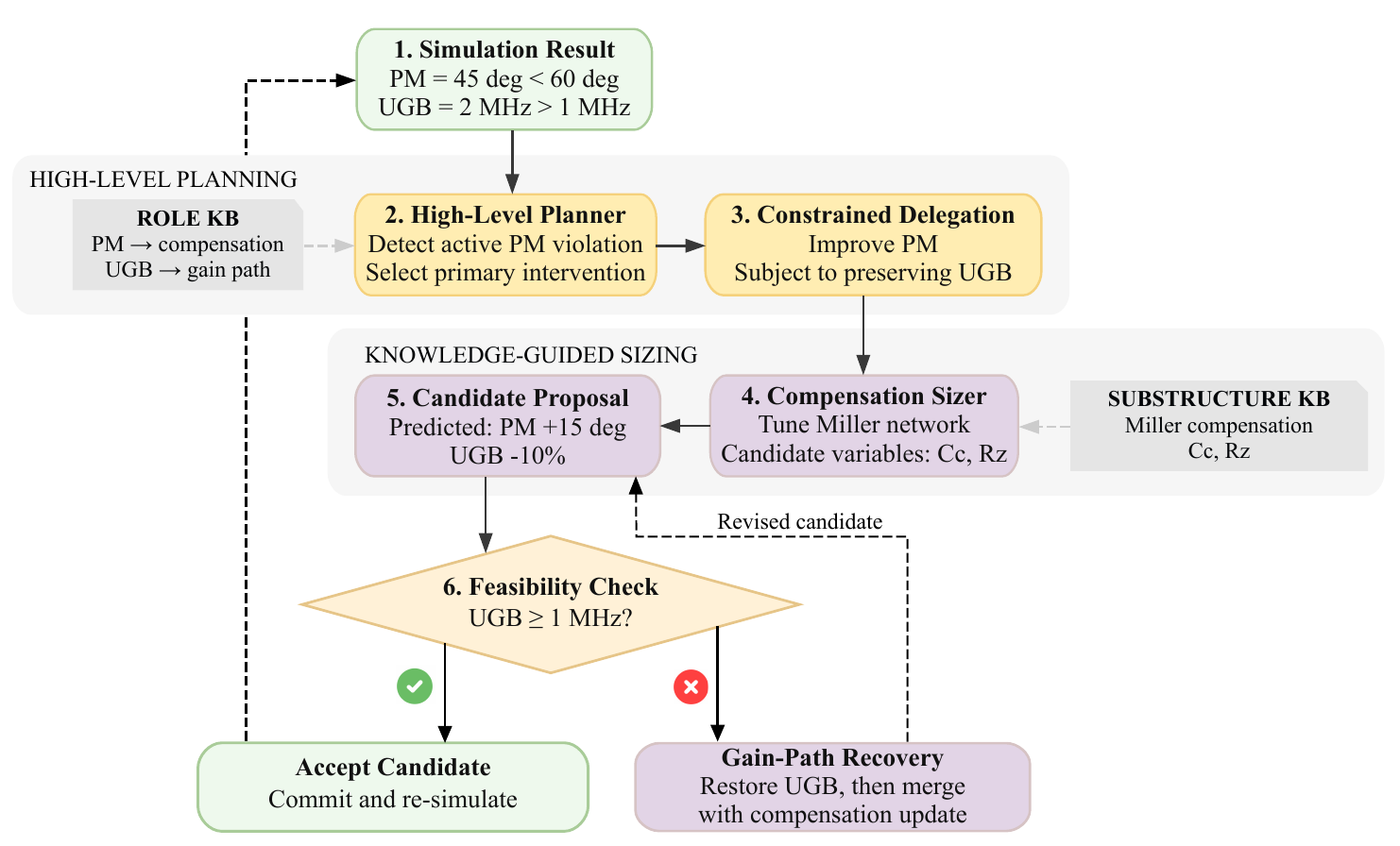}
\caption{The illustration of planner-sizer rounds for Miller-compensated OTA phase margin recovery during iterative sizing.}
\label{flow_ota}
\end{figure}

An example illustrating the flow is shown in Fig. \ref{flow_ota}. The example refers to a Miller-compensated OTA in which the post-simulation phase margin is $45^\circ$, below the required specification of
$60^\circ$, while the unity-gain bandwidth is $2\,\mathrm{MHz}$, exceeding its
$1\,\mathrm{MHz}$ requirement. This means phase margin is in active violation, whereas bandwidth is a satisfied but potentially vulnerable constraint.

The high-level planner first interprets the violation using the role-level knowledge base. It identifies the compensation network as the primary functional
role associated with phase-margin recovery and the gain path as a secondary role that may interact with bandwidth and stability. The planner therefore formulates
a role-scoped task for the compensation sizer: improve phase margin while preserving sufficient unity-gain bandwidth.

The compensation sizer then queries the substructure-level knowledge base and identifies the Miller compensation capacitor $C_c$ and the compensation zero-setting resistor $R_z$ as relevant tuning variables. Based on local knowledge relating compensation parameters to stability metrics, it proposes increasing $C_c$ and adjusting $R_z$. The predicted effect is an increase in
phase margin of approximately $15^\circ$, accompanied by an estimated $10\%$ reduction in unity-gain bandwidth.

The planner subsequently evaluates this proposal against the global performance trade-off knowledge base. If the predicted bandwidth remains above its specification after the compensation update, the planner accepts the proposal and commits the candidate design to simulation. If the same update were predicted to violate the bandwidth constraint, the planner would instead reject the update or dispatch a subsequent task to the gain-path sizer to recover bandwidth.

\subsection{LLM knowledge extractor}
In the proposed framework, the knowledge base enables the planner to reuse prior sizing experience, reason about trade-offs, and allocate tasks to suitable circuit roles. It also constrains the sizers by providing role-relevant parameter-performance knowledge, thereby reducing unnecessary exploration before expensive circuit simulations are invoked.

\begin{figure*}[!t]
\centering
\subfloat[]{%
  \includegraphics[width=0.51\columnwidth]
  {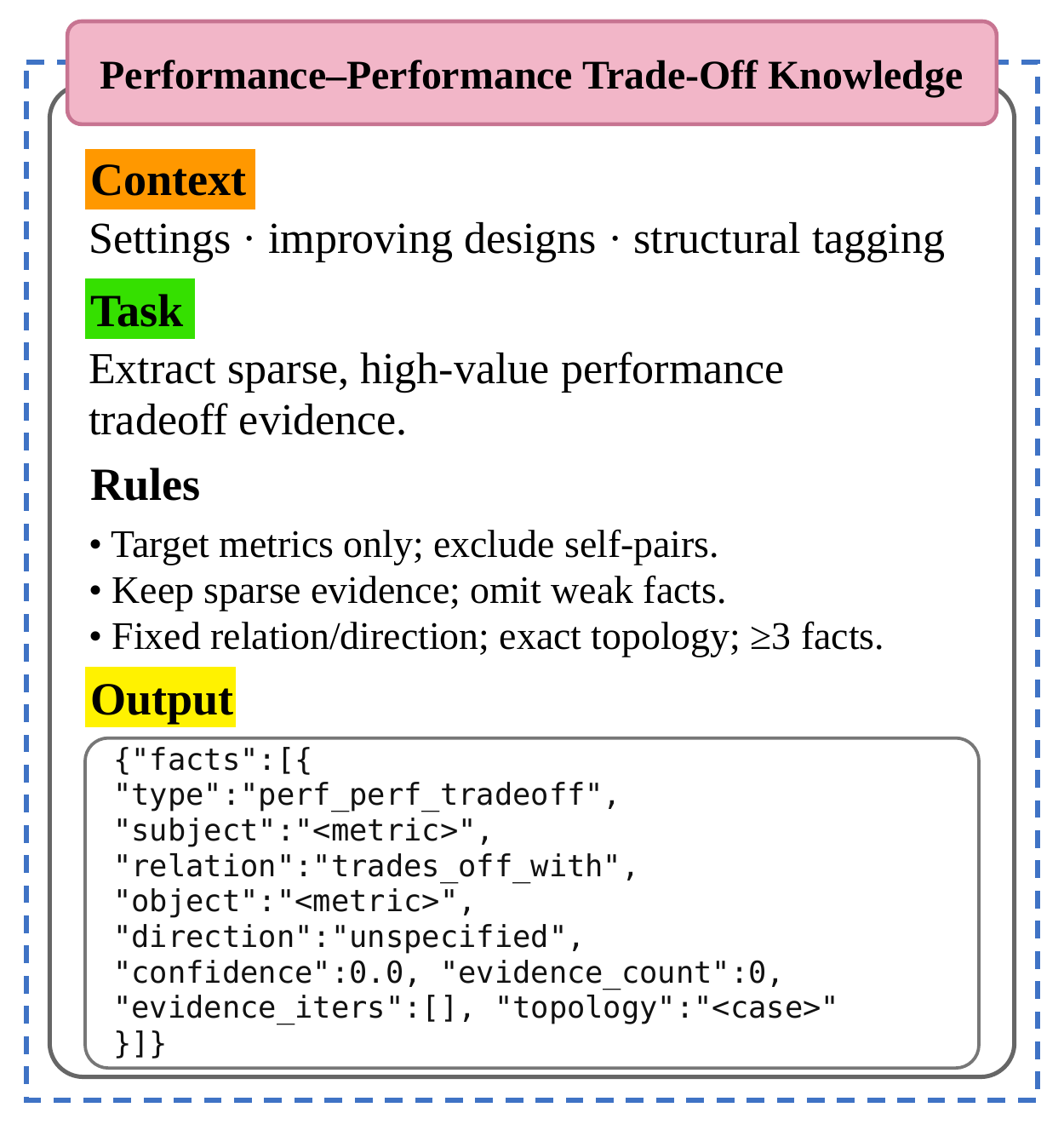}}
\hfill
\subfloat[]{%
  \includegraphics[width=0.51\columnwidth]
  {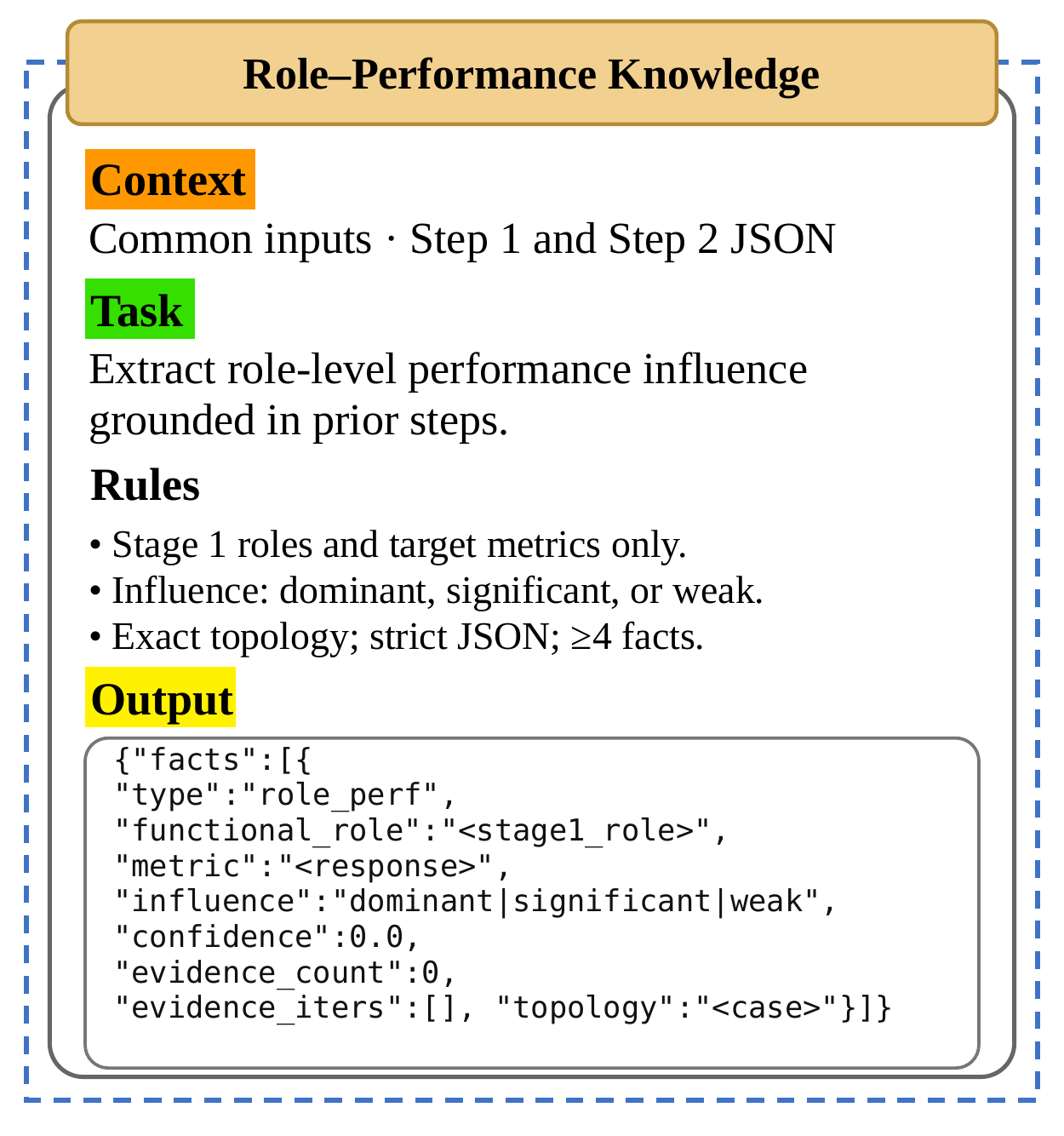}}
\hfill
\subfloat[]{%
  \includegraphics[width=0.51\columnwidth]
  {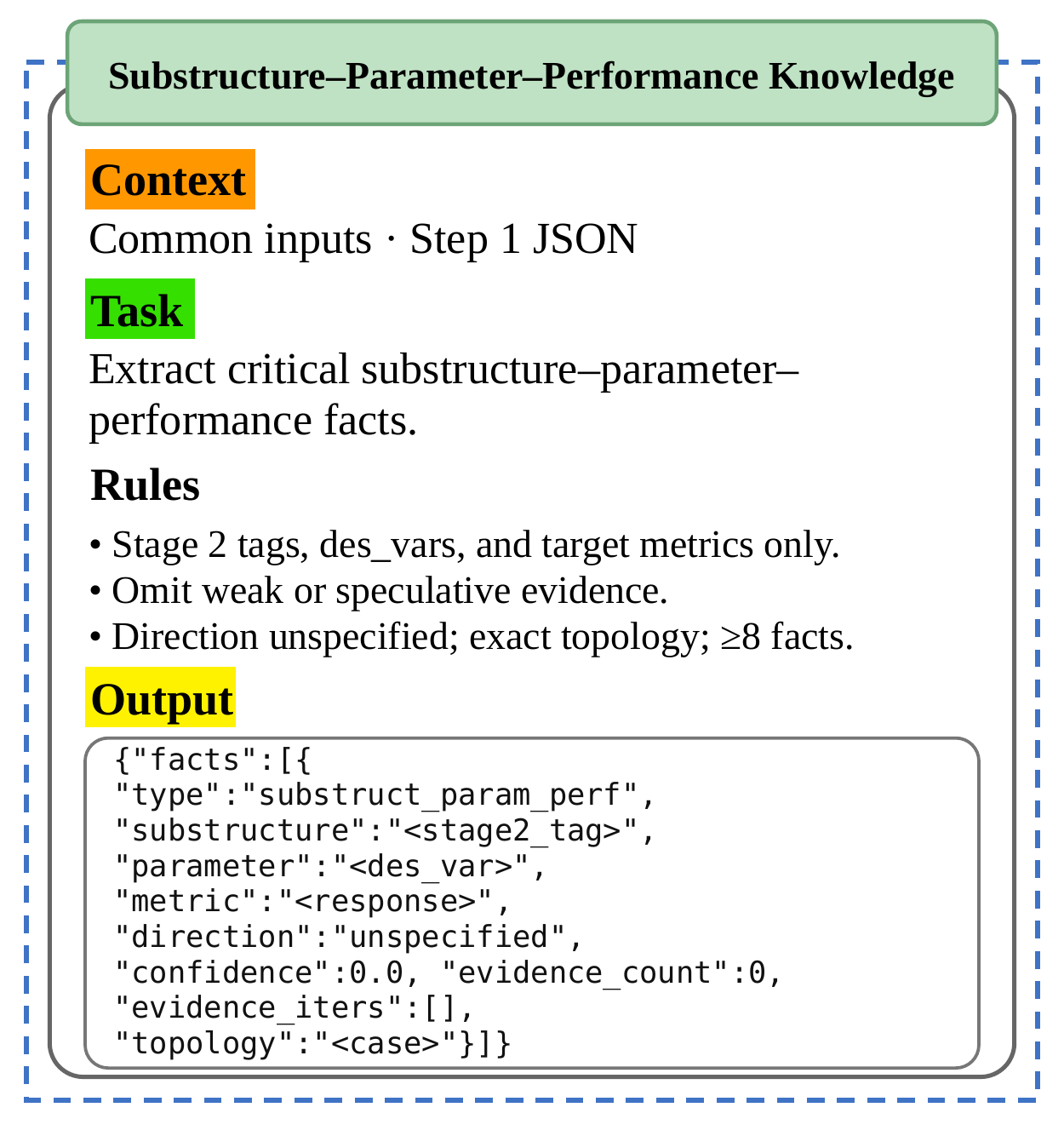}}
\hfill
\subfloat[]{%
  \includegraphics[width=0.51\columnwidth]
  {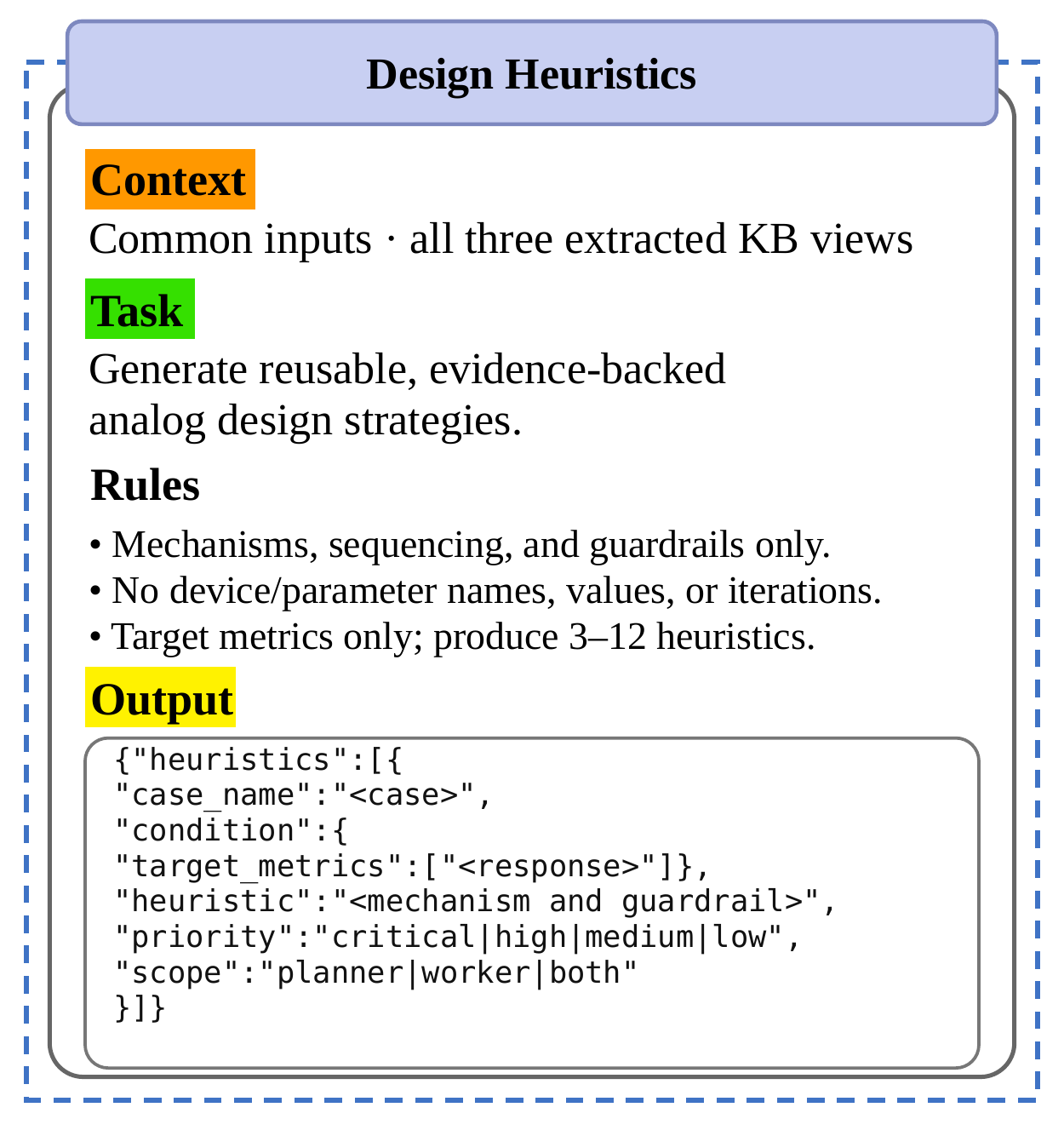}}
\caption{A pipeline that transforms design evidence into four knowledge modules:
(a) performance--performance trade-off knowledge;
(b) role--performance knowledge;
(c) substructure--parameter--performance knowledge; and
(d) design heuristics.}
\label{extractor}
\end{figure*}

The knowledge base organizes analog sizing experience into four structured categories: performance tradeoffs, substructure--parameter--performance relations, role--performance relations, and critical sizing heuristics. Their predefined fields and relation types support automatic validation and task-specific retrieval, allowing the planner to obtain global optimization guidance and the sizers to retrieve knowledge for local parameter updates. Formally, the knowledge base abstracts simulation histories and structural circuit annotations into symbolic facts as shown in Fig. \ref{extractor}.

The first category stores performance-performance trade-off knowledge. It captures relationships among circuit metrics, such as whether improving phase margin tends to reduce unity-gain bandwidth or whether reducing noise tends to increase power consumption. This view is mainly used by the planner to reason about global specification interactions before assigning a sizing task.

The second category stores role-performance knowledge. It associates functional circuit roles, such as the compensation network with the metrics they influence. The influence level may be dominant, significant, or weak. This view supports planner-level task allocation by helping the planner select the most relevant role for a currently violated specification.

The third category stores substructure-parameter-performance knowledge. It links circuit substructures and design variables to performance metrics. For example, in a Miller-compensated two-stage OTA, increasing $C_c$ improves the phase margin, while a separate fact records that increasing $C_c$ reduces the unity-gain bandwidth. This view is primarily used by role-specific sizers to determine which local variables should be modified for a given target metric.

The fourth category stores transferable heuristic knowledge. It captures reusable, evidence-backed design strategies derived from successful sizing trajectories, including general mechanisms, adjustment sequences, and guardrails. Unlike topology-specific knowledge, these heuristics abstract away individual devices, parameter values, and iterations so they can guide future sizing tasks. This view can support both planners and role-specific sizers by providing general design guidance across circuits.

\section{Experiments}
\subsection{Benchmarks}
Table~\ref{tab:benchmarks} summarizes the benchmark suite, organized by circuit complexity. Four textbook circuits \cite{electronics12183833, razavi2017design, gray2009analysis} with fewer than 20 design variables are used to construct the initial design knowledge base and are hence excluded from evaluation. The whole AgenticSizing framework is evaluated on medium-complexity analog blocks, including a two-stage folded-cascode (FC) amplifier \cite{9434374}, bandgap reference (BGR) \cite{5109787}, and low-dropout regulator (LDO) \cite{1233757}, and on a high-complexity LDO integrating both the BGR and FC amplifier.

\begin{figure*}[!ht]
\centering
\subfloat[]{\includegraphics[width=3.1in]{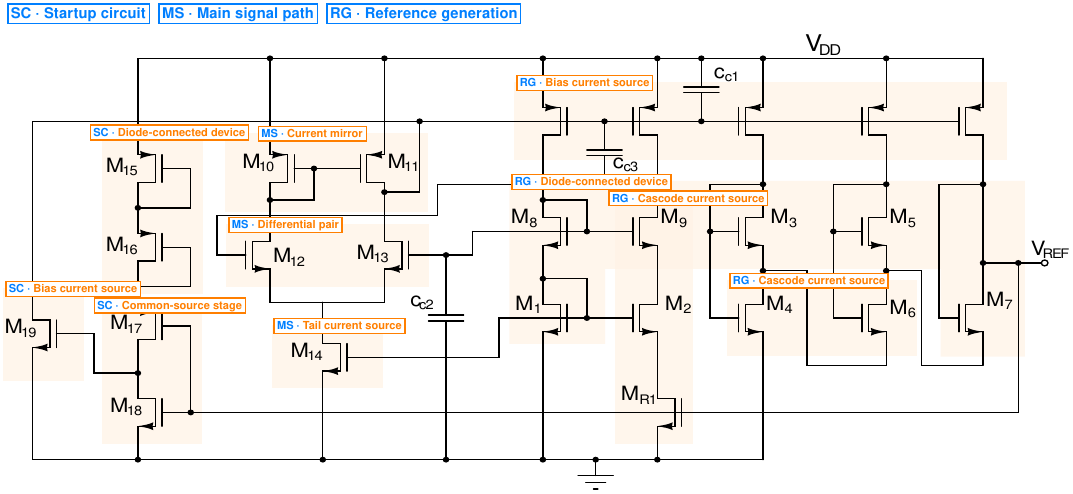}%
\label{fig_first_case}}
\hfil
\subfloat[]{\includegraphics[width=3.8in]{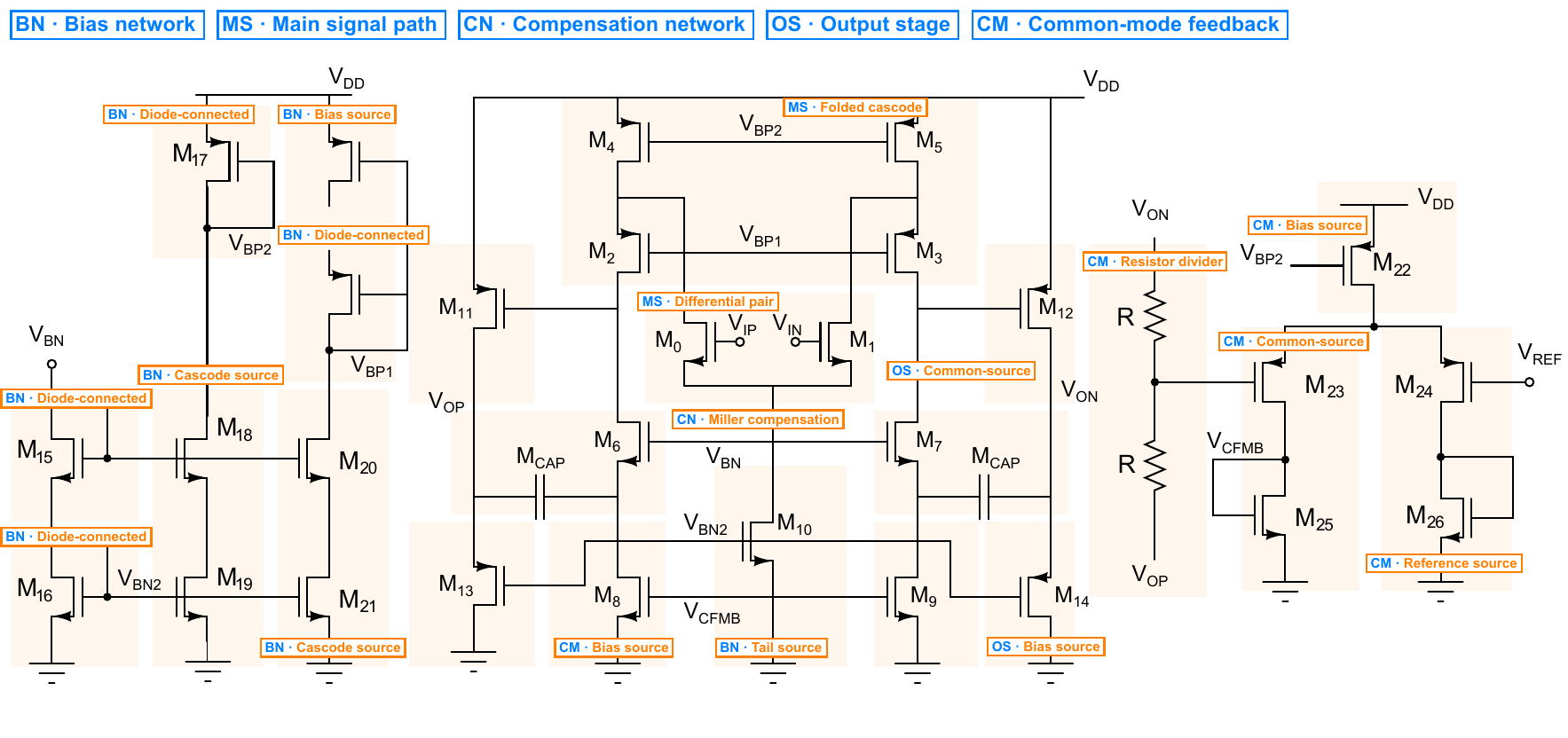}%
\label{fig_second_case}}
\hfil
\subfloat[]{\includegraphics[width=3.8in]{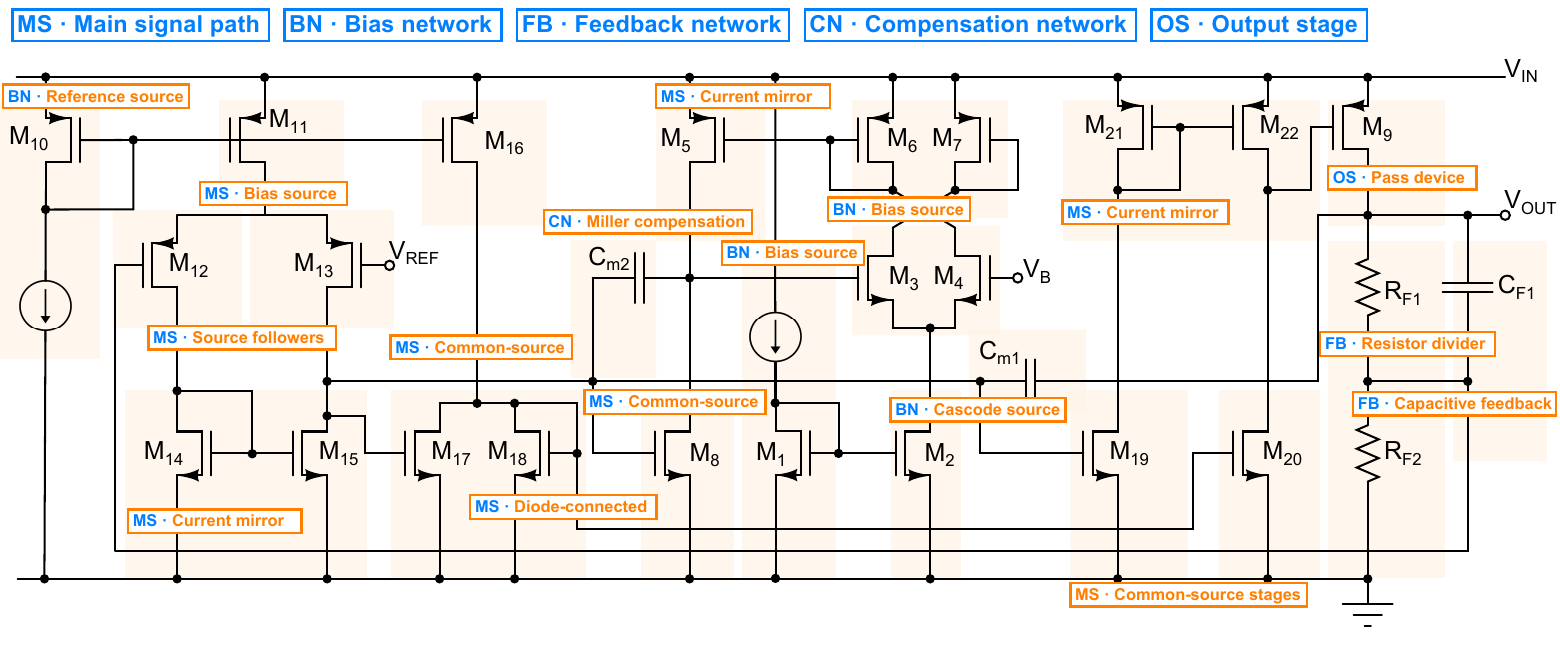}%
\label{fig_third_case}}
\hfil
\subfloat[]{\includegraphics[width=2.5in]{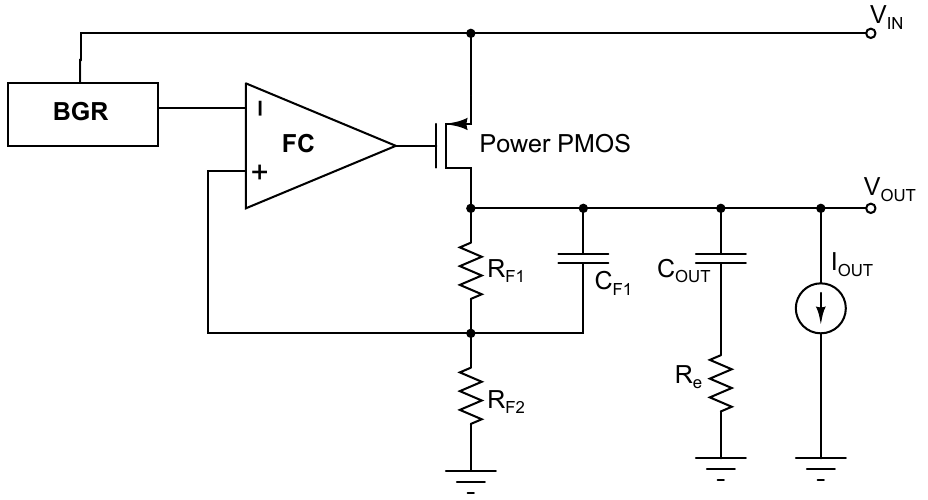}}%
\label{fig_fourth_case}
\hfil
\caption{Topology and tagging results generated by the LLM. The LLM divides circuits into different functional blocks and identifies substructures (coloured) within each block. (a) Bandgap reference circuit. (b) Two-stage folded cascode amplifier. (c) Low dropout regulator. (d) LDO with BGR and two-stage FC amplifier.}
\label{fig_4}
\end{figure*}

Existing LLM-based studies are usually validated only on low- to medium-complexity circuits, leaving limited evidence of their effectiveness on high-complexity designs. To address this gap, we constructed a high-complexity benchmark by integrating medium-complexity blocks into a complete LDO system with a BGR (Fig. \ref{fig_4} (a)) and an FC error amplifier (Fig. \ref{fig_4} (b)). This benchmark contains more than 50 transistors and involves over 50 design variables, shown in Fig. \ref{fig_4} (d).

\begin{table}[t]
\centering
\setlength{\tabcolsep}{3.2pt}
\caption{Summary of benchmarks. The low-complexity circuits are used to construct the initial design knowledge base.}
\label{tab:benchmarks}
\begin{tabular}{l l c c}
\toprule
\textbf{Complexity} & \textbf{Circuit} &
\textbf{\# Variables} & \textbf{\# Transistors} \\
\midrule
\multirow[l]{4}{*}{\makecell[l]{Low\\(initial knowledge\\base source)}}
& 5T-OTA         & 7  & 6  \\
& Telescopic     & 17 & 12 \\
& Current mirror & 10 & 10 \\
& Folded cascode & 16 & 15 \\
\midrule
\multirow{3}{*}{Medium}
& Two-stage FC & 29 & 29 \\
& BGR          & 26 & 25 \\
& LDO          & 26 & 20 \\
\midrule
High
& LDO with BGR and FC & 60 & 55 \\
\bottomrule
\end{tabular}
\end{table}

\subsection{Setup}
Table~\ref{tab:setup_methods} summarizes the setup of the compared methods used in our experiments. We compare the proposed method with three representative optimization baselines: differential evolution (DE), BO, and multi-agent RL (MA-RL). For DE, we adopt the standard DE/rand/1/bin strategy \cite{10.1145/1529255.1529264}. For BO, we use a Gaussian process surrogate model with an RBF kernel and the weighted expected improvement (wEI) acquisition function \cite{8116661}. For MA-RL, we use the MATD3 algorithm with an actor-critic network structure, where the episode depth is set to 40 according to \cite{10525204}. For MA-LLM, we implement with LangGraph and use GPT 5.2 as the backbone language model, with a temperature of 0.5 and a maximum generation length of 5,000 tokens. These settings are kept fixed across all benchmark circuits. The experiments are conducted on a workstation equipped with AMD EPYC 9474F 48-Core Processor and 1.48T memory. The simulator used is Cadence Spectre. Since the compared methods are based on substantially different frameworks, we use wall-clock time as the primary basis for comparison. Each experiment is repeated five times to account for stochastic variability. We compared the performance of our framework with respect to the baselines in terms of success rate, average number of simulations to achieve the best figure of merit (FoM) \cite{8116661}, and wall-clock time. The wall-clock time includes the total optimization time, the optimization time per iteration, and the simulation time per iteration. For the integrated LDO, an additional ablation study evaluates the contributions of topology tagging, multi-agent specialization, and the design knowledge base to optimization success and computational efficiency. Each ablation configuration is evaluated using five independent sizing attempts under the same time budget.

\begin{table}[!t]
\centering
\caption{Setup of compared methods.}
\label{tab:setup_methods}
\renewcommand{\arraystretch}{1.2}
\begin{tabular}{l l l}
\hline
\textbf{Method} & \textbf{Hyper-parameter} & \textbf{Value} \\
\hline
\multirow{1}{*}{DE} & Algorithm & DE/rand/1/bin \\
\hline
\multirow{2}{*}{BO} & Kernel function & RBF \\
& Acquisition function & wEI \\
\hline
\multirow{3}{*}{MA-RL} & Algorithm & MATD3 \\
& Network Structure & Actor--Critic \\
& Episode depth & 40 \\
\hline
\multirow{3}{*}{MA-LLM} & LLM version & GPT 5.2 \\
& Temperature & 0.5 \\
& Max generation token length & 5,000 \\
\hline
\end{tabular}
\end{table}

\subsection{Results on mid-complexity circuits}
\subsubsection{Results on BGR}
The total runtime is set to 4 hours. Table~\ref{tab:design} compares the performance of different optimization approaches on the BGR benchmark across five independent attempts. Overall, DE achieves the highest reliability, satisfying all design specifications in all five attempts. However, DE requires a relatively large number of simulations, with an average of 936 simulations among successful runs. This indicates that while DE is robust for this benchmark, it is simulation-intensive. BO achieves valid designs in three out of five attempts. Compared with DE, BO requires fewer simulations for successful runs, with an average of 208 simulations. However, its robustness is lower, as two attempts fail due to excessive power consumption, and one of these also violates the TC specification. This suggests that BO can be efficient for this circuit, but its performance is sensitive to the search trajectory and surrogate model quality. MARL fails in all five attempts. Although the output reference voltage remains within the target range in all cases, MARL frequently violates the power, PSR, and TC constraints. In particular, several runs produce designs with power consumption far above the target and insufficient PSR. These results indicate that the MA-RL baseline struggles to handle the constrained analog sizing problem for the BGR circuit within the given time budget.

AgenticSizing achieves four successful runs out of five and requires substantially fewer simulations than both DE and BO. Successful AgenticSizing runs find valid designs using only 4 to 45 simulations, with an average of 22.75 simulations. In addition to its simulation efficiency, AgenticSizing also finds high-quality designs, including the lowest power consumption among all successful automated methods, 0.106 $\mu$W, and the strongest PSR, $-90.26$ dB. The only failed AgenticSizing run is close to feasibility, violating only the TC constraint slightly, with 25.72 ppm compared with the 25 ppm target. It is also important to note that the reduction in the number of simulations is achieved at the expense of long LLM reasoning time and significant token consumption, which should be considered during practical application.

The manual design satisfies the voltage, PSR, and TC specifications but slightly exceeds the power constraint, with 1.051 $\mu$W compared with the 1 $\mu$W target. In contrast, AgenticSizing identifies multiple feasible designs with lower power than the manual baseline. These results demonstrate that AgenticSizing provides a favourable balance between success rate, design quality, and simulation efficiency on the BGR benchmark.

\begin{table}[!t]
\centering
\scriptsize
\setlength{\tabcolsep}{6pt}
\caption{BGR: Best performance set found by different approaches (across five attempts). If any specifications are not met within four hours, it is marked in red and counted as a failure.}
\label{tab:design}
\begin{tabular}{l c c c c c}
\toprule
\multirow{2}{*}{Method} & \multirow{2}{*}{Sim.} & Power & Vref & PSR & TC \\
 & & ($\mu$W) & (V) & (dB) & (ppm) \\
\midrule
\textbf{Target} & - & $\le 1$ & $[0.5, 1.3]$ & $\le -40$ & $\le 25$ \\
\midrule
\midrule
\multirow{5}{*}{DE} & 726 & 0.460 & 0.62 & $-$63.6 & 12.26 \\
 & 134 & 0.671 & 0.57 & $-$72.5 & 13.29 \\
 & 1198 & 0.610 & 0.54 & $-$60.3 & 22.74 \\
 & 1440 & 0.260 & 0.65 & $-$46.7 & 15.66 \\
 & 1182 & 0.248 & 0.61 & $-$46.2 & 2.56 \\
\midrule
\multirow{5}{*}{BO}  & 213 & 0.658 & 0.68 & $-$82.1 & 18.24 \\
 & fail & \textcolor{red}{2.810} & 0.71 & $-$76.6 & 23.42 \\
 & 265 & 0.417 & 0.53 & $-$48.6 & 9.82 \\
 & 145 & 0.477 & 0.75 & $-$53.6 & 0.23 \\%
 & fail & \textcolor{red}{9.258} & 0.67 & $-$72.5 & \textcolor{red}{113.32} \\
\midrule
\multirow{5}{*}{MARL} 
 & fail & \textcolor{red}{9.741} & 0.52 & \textcolor{red}{$-$8.7} & \textcolor{red}{261.81} \\
 & fail & \textcolor{red}{833.422} & 0.91 & \textcolor{red}{$-$10.5} & \textcolor{red}{34.54} \\
 & fail & 1.993 & 0.70 & $-$69.6 & \textcolor{red}{26.99} \\
 & fail & \textcolor{red}{47.224} & 0.59 & $-$62.3 & \textcolor{red}{52.17} \\
 & fail & \textcolor{red}{349.211} & 0.71 & \textcolor{red}{$-$12.6} & \textcolor{red}{64.42} \\
\midrule
\multirow{5}{*}{AgenticSizing} 
 & 33 & 0.308 & 0.58 & $-$90.3 & 16.64 \\
 & fail & 0.231 & 0.53 & $-$70.2 & \textcolor{red}{25.72} \\
 & 4 & 0.317 & 0.51 & $-$80.7 & 9.24 \\
 & 45 & 0.106 & 0.61 & $-$69.9 & 19.57  \\
 & 9 & 0.452 & 0.59 & $-$73.8 & 19.89 \\
\midrule
\midrule
\multirow{1}{*}{Manual} & -- & \textcolor{red}{1.051} & 0.71 & $-$89.0 & 9.69 \\
\bottomrule
\end{tabular}
% }
\end{table}

\subsubsection{Results on two-stage FC}
Table \ref{tab:FC} summarizes the results for the two-stage FC sizing task. DE, BO, and MARL fail to satisfy all design specifications within the two-hour computational budget in every trial. The most frequent violations occur in the noise, rise time, UGB, and CMPM constraints. In contrast, AgenticSizing achieves a 80\% success rate. It converges within only 11--29 circuit simulations while satisfying all target specifications. The sole failure case exhibits a power value close to the target. The manual design meets the specified requirements, except for the power requirement. These results demonstrate that AgenticSizing is more reliable and sample-efficient than both conventional optimization and learning-based methods. Similar trends were also observed in other amplifier sizing experiments, suggesting that the LLM-based agent has high performance on analog amplifier sizing tasks.

\begin{table*}[ht!]
\centering
\scriptsize
\setlength{\tabcolsep}{6pt}
\caption{Two-stage FC: Best performance set found by different approaches (across five attempts). If any specifications are not met within two hours, it is marked as a failure.}
\label{tab:FC}
\begin{tabular}{l c c c c c c c c c c c c}
\toprule
\multirow{2}{*}{Method} & \multirow{2}{*}{Sim.} & Power & CMRR & ADM & PSRR & Output swing & PM & Noise & Rise Time & Slew & UGB & CMPM \\
 & & (mW) & (V) & (dB) & (dB) & (V) & ($^\circ$) & ($\mu$V) & (ns) & (V) & (MHz) & ($^\circ$)\\
\midrule
\textbf{Target} & - & $\le 1$ & $\ge 100$ & $\ge 90$ & $\ge 100$ & $\ge 2.4$ & $\ge 60$ & $\le 335$ & $\le 30$ & $\le -0.3$ & $\ge 30$ & $\ge 50$\\
\midrule
\midrule
\multirow{5}{*}{DE}  & fail & 1.543 & 103 & 105 & 132 & 2.69 & 66 & \textcolor{red}{434.1} & \textcolor{red}{45.91} & $-$0.53 & 37.67 & \textcolor{red}{27}
\\
 & fail & 1.281 & 96 & 96 & 126 & 2.86 & 74 & \textcolor{red}{392.5} & \textcolor{red}{42.74} & $-$0.57 & \textcolor{red}{27.51} & \textcolor{red}{15}
\\
 & fail & 1.672 & 95 & 95 & 124 & 3.57 & 82 & \textcolor{red}{365.1} & \textcolor{red}{43.92} & $-$0.51 & \textcolor{red}{24.67} & \textcolor{red}{20} \\
 & fail & 0.962 & 71 & \textcolor{red}{71} & \textcolor{red}{93} & 2.96 & 88 & 289.5 & \textcolor{red}{41.00} & $-$0.51 & \textcolor{red}{27.23} & \textcolor{red}{23} \\
 & fail & 1.403 & 97 & 95 & 106 & 3.51 & 72 & 262.5 & \textcolor{red}{32.71} & $-$0.55 & 43.84 & \textcolor{red}{25}\\
\midrule
\multirow{5}{*}{BO}  & fail & 0.921 & \textcolor{red}{99} & \textcolor{red}{87} & \textcolor{red}{89} & 2.94 & 68 & 241.2 & \textcolor{red}{47.69} & $-$0.67 & \textcolor{red}{29.76} & 62 \\
 & fail & 0.966 & 104 & 105 & 143 & 2.81 & 84 & \textcolor{red}{365.9} & \textcolor{red}{73.79} & $-$0.56 & \textcolor{red}{21.85} & \textcolor{red}{29} \\
 & fail & 0.676 & 110 & 110 & 134 & 2.57 & 78 & \textcolor{red}{348.1} & \textcolor{red}{56.63} & $-$0.65 & \textcolor{red}{22.64} & 61\\
 & fail & \textcolor{red}{1.283} & 101 & 97 & 104 & 3.53 & 76 & \textcolor{red}{430.8} & 27.53 & $-$0.55 & \textcolor{red}{25.77} & \textcolor{red}{42} \\
 & fail & 0.974 & 114 & 114 & 154 & 2.83 & \textcolor{red}{47} & 328.4 & \textcolor{red}{79.08} & $-$0.69 & \textcolor{red}{28.70} & \textcolor{red}{17} \\
\midrule
\multirow{5}{*}{MARL}  
 & fail & \textcolor{red}{1.185} & 100 & 100 & 125 & 3.51 & 91 & \textcolor{red}{472.6} & \textcolor{red}{59.63} & $-$0.48 & \textcolor{red}{27.30} & \textcolor{red}{25} \\
 & fail & 0.533 & 88 & 92 & 138 & 2.73 & \textcolor{red}{27} & \textcolor{red}{393.4} & \textcolor{red}{116.6} & $-$0.46 & 41.88 & \textcolor{red}{29} \\
 & fail & \textcolor{red}{1.250} & 98 & 98 & 123 & 3.31 & 91 & \textcolor{red}{444.4} & \textcolor{red}{64.56} & $-$0.45 & \textcolor{red}{25.52} & \textcolor{red}{10} \\
 & fail & \textcolor{red}{1.490} & 90 & 90 & 131 & 3.56 & 76 & \textcolor{red}{427.2} & \textcolor{red}{35.37} & $-$0.52 & \textcolor{red}{20.65} & \textcolor{red}{17} \\
 & fail & \textcolor{red}{1.403} & 101 & 101 & 127 & 3.48 & 70 & \textcolor{red}{356.8} & \textcolor{red}{52.02} & $-$0.37 & \textcolor{red}{17.64} & \textcolor{red}{12} \\
\midrule
\multirow{5}{*}{AgenticSizing} & 13 & 0.975 & 109 & 108 & 123 & 2.81 & 88 & 306.1 & 29.90 & $-$0.53 & 56 & 51 \\ 
 & 11 & 0.788 & 105 & 100 & 106 & 3.28 & 92 & 332.8 & 29.95 & $-$0.56 & 349 & 63 \\ 
 & 29 & 0.784 & 103 & 100 & 111 & 2.82 & 79 & 334.5 & 29.98 & $-$0.56 & 217 & 61 \\
 & fail & \textcolor{red}{1.040} & 116 & 113 & 120 & 2.74 & 86 & 329.6 & 29.92 & $-$0.57 & 272 & 52 \\
 & 24 & 0.946 & 109 & 109 & 130 & 2.79 & 80 & 313.6 & 28.37 & $-$0.58 & 51.47 & 50 \\
\midrule
\midrule
\multirow{1}{*}{Manual} & - & \textcolor{red}{1.272} & 107 & 109 & 131 & 3.11 & 73 & 332.6 & 26.39 & $-$0.57 & 37.58 & 52 \\
\bottomrule
\end{tabular}
% }
\end{table*}

\subsubsection{Results on LDO}
Table~\ref{tab:ldo} summarizes the LDO sizing results over five independent trials within a four-hour budget. BO and AgenticSizing achieve comparable reliability with a 4/5 success rate, whereas DE and MARL succeed in only 1/5 trials. However, AgenticSizing demonstrates significantly higher simulation efficiency, converging within 64--103 simulations (79 on average for successful runs), compared with approximately 524 simulations for BO and 1131 simulations for the successful DE trial.

The results reveal different optimization characteristics. DE satisfies the regulation and PSRR constraints, but often fails to achieve the target output voltage, indicating difficulty in balancing the coupled specifications. BO provides robust convergence but requires substantially more simulations, reflecting the cost of black-box exploration. MARL exhibits poor robustness. This behavior is likely attributed to the challenges of RL in analog sizing, where sparse reward feedback and costly environment interactions make it difficult to efficiently learn a policy for highly constrained design spaces. In contrast, AgenticSizing achieves high reliability with minimal simulation overhead. This efficiency is attributed to the LLM's in-context learning capability and incorporation of prior domain-specific knowledge, which enable more informed design exploration and reduce unnecessary simulations. Its only failed trial violates the power constraint by 5.6\% while satisfying all other specifications. The manual design meets all specifications with a power consumption of 81~\textmu W, which is slightly higher than the 33--74.5~\textmu W achieved by AgenticSizing.

\begin{table}[ht!]
\centering
\scriptsize
\setlength{\tabcolsep}{2pt}
\caption{LDO: Best performance set found by different approaches across five attempts. If any specifications are not met within four hours, it is marked as a failure.}
\label{tab:ldo}
\begin{tabular}{l c c c c c c}
\toprule
\multirow{2}{*}{Method} & \multirow{2}{*}{Sim.} 
& Power & Vout & Load Regulation & Line Regulation & PSRR \\
& & (\textmu W) & (V) & (V/A) & (V/V) & (dB) \\
\midrule
\textbf{Target} & -- & $\le 100$ & $[1.15, 1.25]$ & $\le 0.2$ & $\le 0.2$ & $\ge 50$ \\
\midrule
\midrule
\multirow{5}{*}{DE} 
& fail & 51.1 & \textcolor{red}{1.38} & 0.045 & 0.0003 & 68.0 \\
& fail & 198.9 & \textcolor{red}{1.35} & 0.064 & 0.0018 & 62.9 \\
& 1131 & 27.4 & 1.17 & 0.090 & 0.0007 & 73.5 \\
& fail & 120.7 & \textcolor{red}{1.40} & 0.044 & 0.0007 & 62.9 \\
& fail & 80.5 & \textcolor{red}{1.37} & \textcolor{red}{0.202} & 0.0016 & 55.6 \\
\midrule
\multirow{5}{*}{BO}  
& 475 & 25.9 & 1.25 & 0.027 & 0.0004 & 68.0 \\
& fail & \textcolor{red}{762.1} & \textcolor{red}{1.66} & \textcolor{red}{2.456} & 0.1682 & 50.6 \\
& 563 & 35.9 & 1.22 & 0.035 & 0.0004 & 68.7 \\
& 533 & 38.7 & 1.23 & 0.033 & 0.0007 & 62.6 \\
& 526 & 17.1 & 1.15 & 0.066 & 0.0024 & 69.6 \\
\midrule
\multirow{5}{*}{MARL} 
& fail & 69.1 & \textcolor{red}{0.81} & \textcolor{red}{90.613} & 0.0047 & \textcolor{red}{40.2} \\
& 542 & 57.2 & 1.23 & 0.040 & 0.0005 & 65.2 \\
& fail & \textcolor{red}{127.0} & \textcolor{red}{0.95} & \textcolor{red}{92.051} & 0.0278 & \textcolor{red}{29.7} \\
& fail & 18.7 & \textcolor{red}{0.93} & 0.172 & 0.0126 & \textcolor{red}{37.6} \\
& fail & 17.2 & \textcolor{red}{0.92} & \textcolor{red}{0.211} & 0.0003 & 72.8 \\
\midrule
\multirow{5}{*}{AgenticSizing} 
& 77 & 33.0 & 1.16 & 0.050 & 0.0021 & 54.5 \\
& 64 & 74.5 & 1.20 & 0.026 & 0.0062 & 51.2 \\
& fail & \textcolor{red}{105.6} & 1.16 & 0.067 & 0.0045 & 54.0 \\
& 73 & 44.2 & 1.15 & 0.026 & 0.0002 & 75.2 \\
& 103 & 57.3 & 1.22 & 0.055 & 0.0021 & 53.9 \\
\midrule
\midrule
Manual & - & 81 & 1.20 & 0.12 & 0.0005 & 55.3 \\
\bottomrule
\end{tabular}
\end{table}

\subsection{Full analysis on LDO with FC and BGR (with ablation study)}
Table~\ref{tab:ldo_full} reports the full-LDO sizing results under a six-hour budget. DE, BO, and MARL fail to satisfy all specifications in every attempt, with failures primarily caused by output-voltage, load-regulation, and PSRR violations. MARL exhibits the largest performance variability, including severe load-regulation and temperature-coefficient violations, highlighting the difficulty of RL in navigating the highly constrained analog design space with sparse performance feedback. In contrast, AgenticSizing achieves the highest success rate, identifying feasible designs in three out of five trials within 64--94 simulations, demonstrating superior robustness and sample efficiency. The failures are limited to output-voltage and PSRR violations, suggesting that the full-LDO sizing problem presents substantially tighter and more strongly coupled design constraints than the standalone LDO benchmark. These results indicate that leveraging LLM in-context learning together with domain-specific knowledge enables more effective exploration of the constrained design space than conventional black-box or RL-based optimization approaches.

\newcommand{\bad}[1]{\textcolor{red}{#1}}
\begin{table}[!t]
\centering
\scriptsize
\renewcommand{\arraystretch}{1.15}
\setlength{\tabcolsep}{3pt}

\caption{Full LDO: Performance sets found by different approaches across five attempts. If any specification is not met within six hours, the attempt is marked as a failure.}
\label{tab:ldo_full}

\begin{tabular}{l c c c c c c c}
\toprule
\multirow{2}{*}{Method} 
& \multirow{2}{*}{Sim.} 
& Power 
& Vout 
& \makecell{Load\\Reg.} 
& \makecell{Line\\Reg.} 
& TC
& PSRR \\
& 
& (\textmu W) 
& (V) 
& (V/A) 
& (V/V) 
& (ppm)
& (dB) \\
\midrule

\textbf{Target} 
& -- 
& $\le 700$ 
& $[1.15, 1.25]$ 
& $\le 0.2$ 
& $\le 0.2$ 
& $\le 40$  
& $\ge 50$ \\

\midrule
\midrule

\multirow{5}{*}{DE} 
& fail & 863.8 & 1.19 & 0.178 & 0.0047 & 41.7 & \textcolor{red}{43.8} \\
& fail & 642.4 & \textcolor{red}{0.83} & \textcolor{red}{0.532} & 0.0381 & 29.3 & \textcolor{red}{27.4} \\
& fail & \textcolor{red}{1408.1} & \textcolor{red}{1.34} & 0.169 & 0.0287 & 24.1 & 74.7 \\
& fail & 695.2 & \textcolor{red}{0.87} & \textcolor{red}{0.297} & 0.0196 & 36.0 & \textcolor{red}{29.5} \\
& fail & 918.5 & \textcolor{red}{1.03} & \textcolor{red}{0.567} & 0.0349 & 39.8 & \textcolor{red}{27.5} \\
\midrule

\multirow{5}{*}{BO}  
& fail & 739.8 & 1.17 & \textcolor{red}{0.208} & 0.0236 & 24.7 & \textcolor{red}{31.3} \\
& fail & 968.9 & \textcolor{red}{1.01} & 0.196 & 0.0066 & 24.0 & \textcolor{red}{39.0} \\
& fail & 895.0 & \textcolor{red}{1.12} & 0.172 & 0.0054 & 33.0 & \textcolor{red}{40.3} \\
& fail & 290.7 & \textcolor{red}{0.77} & 0.023 & 0.0028 & 32.2 & 50.0 \\
& fail & 821.2 & \textcolor{red}{0.93} & 0.187 & 0.0209 & 31.5 & \textcolor{red}{30.9} \\
\midrule

\multirow{5}{*}{MARL} 
& fail & 867.8 & \textcolor{red}{0.77} & \textcolor{red}{0.367} & 0.0158 & \textcolor{red}{163.9} & \textcolor{red}{35.6} \\
& fail & 823.8 & 1.24 & \textcolor{red}{0.801} & 0.0337 & \textcolor{red}{48.4} & \textcolor{red}{28.9} \\
& fail & 214.1 & \textcolor{red}{0.65} & \textcolor{red}{3.025} & 0.1391 & \textcolor{red}{245.0} & \textcolor{red}{15.3} \\
& fail & 486.6 & \textcolor{red}{0.61} & 0.051 & 0.0024 & 11.3 & \textcolor{red}{45.6} \\
& fail & 776.9 & \textcolor{red}{1.37} & \textcolor{red}{0.901} & 0.0221 & \textcolor{red}{44.0} & \textcolor{red}{26.6} \\
\midrule

\multirow{5}{*}{AgenticSizing} 
& 94 & 591.4 & 1.22 & 0.084 & 0.0028 & 28.8 & 58.3 \\
& 83 & 748.0 & 1.15 & 0.161 & 0.0090 & 39.7 & 65.5 \\ 
& fail & 772.8 & \textcolor{red}{1.11} & \textcolor{red}{0.343} & 0.0610 & 23.8 & \textcolor{red}{22.9} \\
& 92 & 984.7 & 1.21 & 0.171 & 0.0022 & 30.9 & 50.4 \\
& fail & 924.8 & \textcolor{red}{1.14} & 0.082 & 0.0015 & 34.7 & 51.7 \\
\bottomrule
\end{tabular}
\end{table}

\begin{figure*}[t!]
\centering
\includegraphics[width=0.88\textwidth]
  {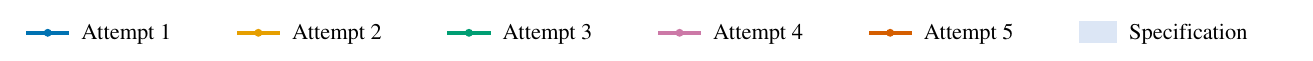}

\subfloat[]{
  \includegraphics[width=0.235\textwidth]
  {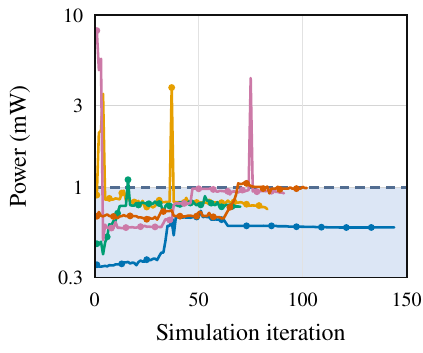}}
\hfill
\subfloat[]{
  \includegraphics[width=0.235\textwidth]
  {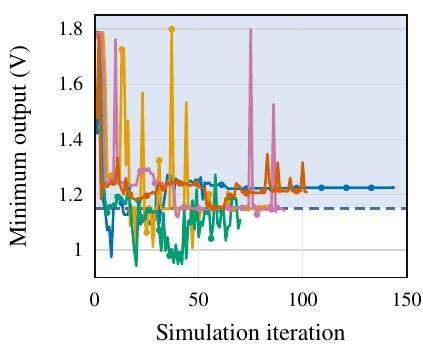}}
\hfill
\subfloat[]{
  \includegraphics[width=0.235\textwidth]
  {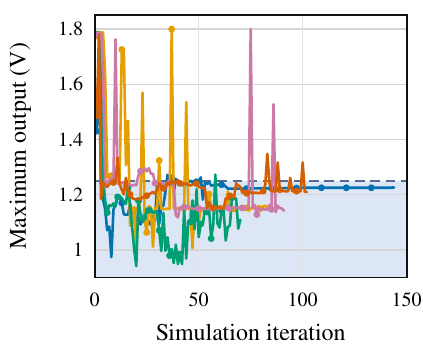}}
\hfill
\subfloat[]{
  \includegraphics[width=0.235\textwidth]
  {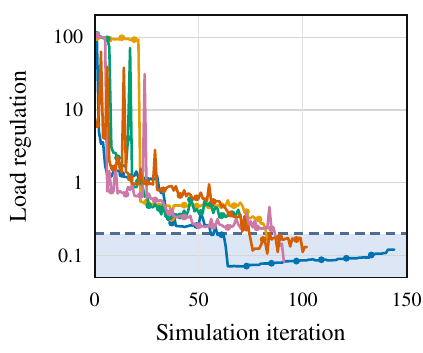}}

\medskip

\hfill
\subfloat[]{
  \includegraphics[width=0.235\textwidth]
  {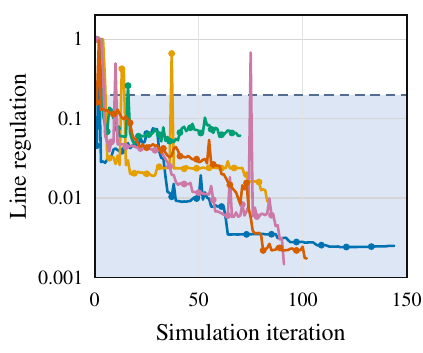}}
\hfill
\subfloat[]{
  \includegraphics[width=0.235\textwidth]
  {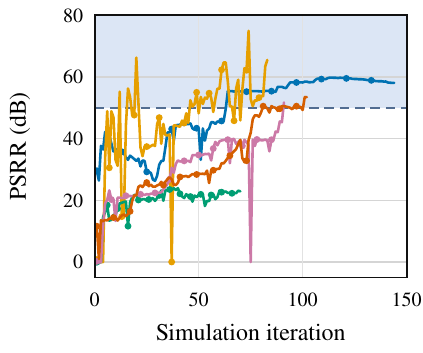}}
\hfill
\subfloat[]{
  \includegraphics[width=0.235\textwidth]
  {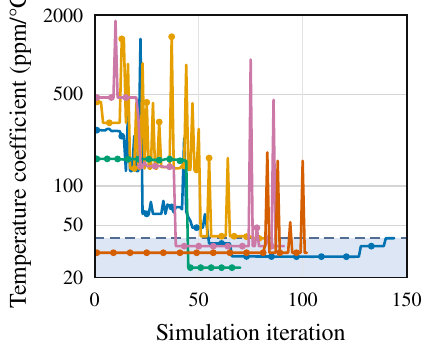}}
\hfill

\caption{Optimization trajectories of performance metrics over five independent attempts for the integrated LDO with FC and BGR. Three attempts reached a specification-compliant design, while two remained infeasible within the time budget. Shaded regions denote specification-compliant ranges.}
\label{fig:optimization_trend}
\end{figure*}

Fig. \ref{fig:optimization_trend} shows the optimization trajectories of five independent runs. Attempts 1, 2, and 5 first achieved feasibility at iterations 64, 83, and 81, respectively, while Attempts 3 and 4 remained infeasible. Attempt 3 rapidly approached the feasible region, but subsequently oscillated near the specification boundaries, particularly for minimum output voltage, load regulation, and PSRR. This limit-cycle-like behavior reflects coupled nonlinear trade-offs and the reduced reliability of qualitative LLM reasoning for fine-grained convergence. Overall, the results demonstrate effective multi-specification coordination while highlighting the run-to-run variability of LLM-guided optimization.

\subsubsection{Efficency analysis across complexity levels}
Table~\ref{tab:efficiency} summarizes the computational efficiency of all methods across benchmarks of increasing complexity, from standalone analog building blocks to fully integrated LDO with FC and BGR system. A clear trend emerges: as circuit complexity and cross-block coupling increase, the robustness and sample efficiency of conventional optimization methods deteriorate rapidly, whereas AgenticSizing consistently maintains competitive success rates with substantially fewer circuit simulations.

Among the baseline methods, DE is effective for relatively simple problems but requires a large number of simulations and fails to scale to more constrained benchmarks. BO generally provides higher robustness but at the cost of extensive black-box exploration, resulting in significantly higher simulation counts. MARL exhibits the weakest scalability, reflecting the difficulty of learning effective policies from sparse reward signals in expensive analog design environments.

In contrast, AgenticSizing consistently requires one to two orders of magnitude fewer simulations while maintaining comparable or higher success rates. A common characteristic of LLM-based methods is that each optimization iteration incurs additional reasoning overhead. This overhead becomes less pronounced as system complexity increases and SPICE simulation time dominates the total runtime. By leveraging in-context learning together with domain-specific knowledge, AgenticSizing efficiently guides the search toward promising design regions, reducing unnecessary circuit evaluations. The advantage becomes increasingly pronounced as the design problem grows in dimensionality and constraint coupling, where AgenticSizing remains the only method capable of finding feasible solutions for the integrated benchmark.

Overall, these results demonstrate that AgenticSizing manages good trade-off between robustness, scalability, and sample efficiency, making it promising for handling complex analog circuit sizing tasks.

\begin{table}[!t]
\centering
\scriptsize
\setlength{\tabcolsep}{2pt}
\caption{Computation efficiency comparison between different approaches averaged across five attempts.}
\label{tab:efficiency}
\begin{tabular}{clccccc}
\toprule
Benchmark &
Method &
\makecell{Succ.\\Rate} &
\makecell{Avg.\\Iter.} &
\makecell{Total\\Opt. Time} &
\makecell{Opt. Time\\per Iter.} &
\makecell{Sim. Time\\per Iter.} \\
\midrule

\multirow{4}{*}{BGR}
& DE            & 5/5 & 936 & 392\,ms & 15.29\,ms & 9\,min\,45\,s \\
& BO            & 3/5 & 208 & 1\,h\,40\,min\,11\,s & 16.25\,s & 24.26\,s \\
& MARL          & 0/5 & --  & 2\,min\,49\,s & 0.30\,s & 24.72\,s \\
& AgenticSizing & 4/5 & 23  & 1\,h\,11\,min\,2\,s & 121.91\,s & 36.83\,s \\
\midrule

\multirow{4}{*}{\makecell{Two-stage\\FC}}
& DE            & 0/5 & -- & 74\,ms & 10.63\,ms & 17\,min\,45\,s \\
& BO            & 0/5 & -- & 52\,min\,32\,s & 19.34\,s & 15.73\,s \\
& MARL          & 0/5 & --  & 4\,min\,4\,s & 0.34\,s & 19.69\,s \\
& AgenticSizing & 4/5 & 19  & 1\,h\,37\,min\,42\,s & 186.53\,s & 22.18\,s \\
\midrule

\multirow{4}{*}{LDO}
& DE            & 1/5 & 1131 & 97\,ms & 9.48\,ms & 25\,min\,44\,s \\
& BO            & 4/5 & 524  & 1\,h\,18\,min\,22\,s & 10.38\,s & 18.70\,s \\
& MARL          & 1/5 & 542  & 2\,min\,9\,s & 0.23\,s & 24.53\,s \\
& AgenticSizing & 4/5 & 79   & 2\,h\,37\,min\,13\,s & 52.41\,s & 27.70\,s \\
\midrule

\multirow{4}{*}{\makecell{LDO with\\FC and BGR}}
& DE            & 0/5 & -- & 192\,ms & 24.04\,ms & 48\,min\,7\,s \\
& BO            & 0/5 & -- & 2\,h\,38\,min\,30\,s & 26.79\,s & 19.93\,s \\
& MARL          & 0/5 & -- & 11\,min\,26\,s & 0.70\,s & 20.85\,s \\
& AgenticSizing & 3/5 & 90 & 4\,h\,59\,min\,15\,s & 162.68\,s & 31.27\,s \\

\bottomrule
\end{tabular}
\end{table}

\subsubsection{The impact of task decomposition}

\begin{table}[!t]
\centering
\setlength{\tabcolsep}{3pt}
\caption{Ablation study of the proposed framework on LDO with FC and BGR and computation efficiency comparison (averaged across five attempts).}
\label{tab:ablation}
\resizebox{0.48\textwidth}{!}{%
\begin{tabular}{l c c c c c c c}
\toprule
\multirow{2}{*}{Methods} & \multicolumn{3}{c}{Components} & \multicolumn{4}{c}{Metrics} \\
\cmidrule(lr){2-4} \cmidrule(lr){5-8}
 & \makecell{Topology\\Tagging} & \makecell{Multi-Agent\\Sizing} & \makecell{Design\\Knowledge Base} & \makecell{Success\\Rate} & \makecell{Avg.\\Iter.} & \makecell{Tokens to\\feasibility} & \makecell{Total\\Tokens} \\
\midrule
S0 & \ding{55} & \ding{55} & \ding{55} & 0\% & -- & -- & 4.064M \\
S1 & \ding{51} & \ding{55} & \ding{55} & 40\% & 144 & 2.969M & 3.579M \\
S2 & \ding{51} & \ding{51} & \ding{55} & 0\% & -- & -- & 3.370M \\
S3 & \ding{51} & \ding{55} & \ding{51} & 20\% & 136 & 3.221M & 3.387M \\
S4 & \ding{51} & \ding{51} & \ding{51} & 60\% & 90 & 2.771M & 3.714M \\
\bottomrule
\end{tabular}%
}
\end{table}

The effectiveness of topology understanding is evaluated by comparing S0 and S1 in Table \ref{tab:ablation}, both of which use the same single-agent sizing configuration without a design knowledge base and differ only in the use of topology tagging. Without topology tagging, S0 fails in all five attempts, whereas S1 achieves a 40\% success rate. Moreover, topology tagging reduces the average total token consumption from 4.064 million to 3.579 million, corresponding to an 11.9\% reduction. Although the absolute token usage remains nontrivial for this complex circuit, the results indicate that topology tagging enables more targeted parameter updates and improves both sizing effectiveness and computational efficiency.

\subsubsection{The impact of multi-agent sizing framework}
The effect of agent specialization can be seen from comparisons S1–S2 and S3–S4 in Table \ref{tab:ablation}. Without design knowledge, replacing the single agent with specialized sizing agents reduces the success rate from 40\% to 0\%, despite a 5.8\% reduction in total token consumption. This result suggests that task decomposition alone may hinder the coordination of coupled subblocks. In contrast, with the knowledge base enabled, agent specialization increases the success rate from 20\% to 60\%. Among successful runs, it reduces the average iterations to feasibility from 136 to 90 and the tokens to feasibility from 3.221 million to 2.771 million, corresponding to reductions of 33.8\% and 14.0\%, respectively. Although the average total token consumption increases by 9.7\%, the improved success rate and lower cost to feasibility indicate that specialized agents become effective when supported by appropriate design knowledge.  

\subsubsection{The impact of knowledge infusion}
From Table \ref{tab:ablation}, the impact of knowledge infusion can be seen with comparisons S1–S3 and S2–S4. In the single-agent configuration, adding the knowledge base reduces the observed success rate from 40\% to 20\% and increases the tokens to feasibility by 8.5\%, indicating that knowledge injection alone does not ensure improved optimization. In the multi-agent configuration, however, knowledge infusion increases the success rate from 0\% to 60\%, with a 10.2\% increase in average total token consumption. 

Overall, these results demonstrate a complementary interaction between design knowledge and agent specialization: the knowledge base provides role-specific guidance, while the multi-agent decomposition enables such guidance to be applied systematically to individual functional blocks.

\section{Conclusion}
This work presented a multi-agent LLM-based framework for complex analog circuit sizing. The proposed workflow integrates topology understanding, design-knowledge extraction, and role-specialized agents coordinated by a planner. By decomposing circuits into functional blocks and incorporating performance trade-offs and parameter–performance dependencies, the framework enables targeted, interpretable optimization with global specification awareness. Ablation studies demonstrate that topology understanding improves success rates, while agent specialization is most effective when supported by design knowledge. The results show that the framework’s performance arises from the coordinated interaction of structural understanding, knowledge-guided reasoning, and multi-agent collaboration. Compared with black-box and single LLM-based methods, the proposed approach offers a more transparent and sample-efficient solution for high-dimensional analog sizing. Future work will extend the framework to larger analog and mixed-signal systems and advanced technology nodes, while improving inter-agent communication, memory, verification, and knowledge transfer across circuit families and processes.

\section*{Acknowledgments}
We thank EDINA at the University of Edinburgh for supporting the access to the OpenAI LLMs through the ELM gateway. Specifically, GPT-5.5 was used during this work to support the development of the AgenticSizing framework, as well as for manuscript proofreading and language refinement.

\bibliographystyle{IEEEtran}
\bibliography{analog_sizing}

\begin{IEEEbiography}
[{\includegraphics[width=1in,height=1.25in,clip,keepaspectratio]{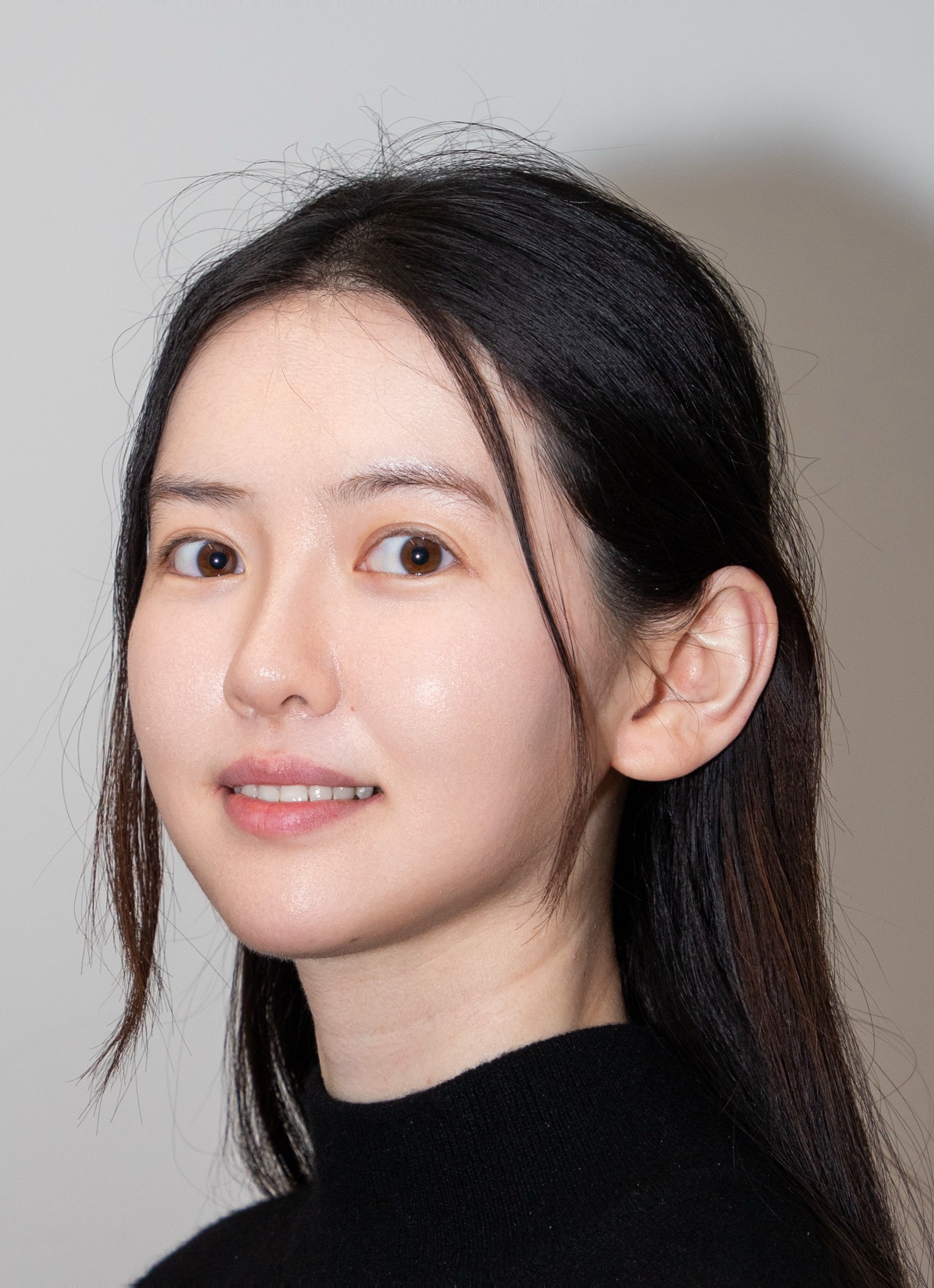}}]{Yijia Hao} received the B.S. degree in electronic and electrical engineering from the University of Electronic Science and Technology of China, Chengdu, China, in 2021. She received the Ph.D. degree with the University of Glasgow, Glasgow, U.K. in 2025.
She worked as a Postdoctoral Researcher with the Centre for Electronics Frontiers, University of Edinburgh, U.K. in 2026. Her current research interest includes analog and mixed-signal IC design. 
    
\end{IEEEbiography}

\begin{IEEEbiography}
[{\includegraphics[width=1in,height=1.25in,clip,keepaspectratio]{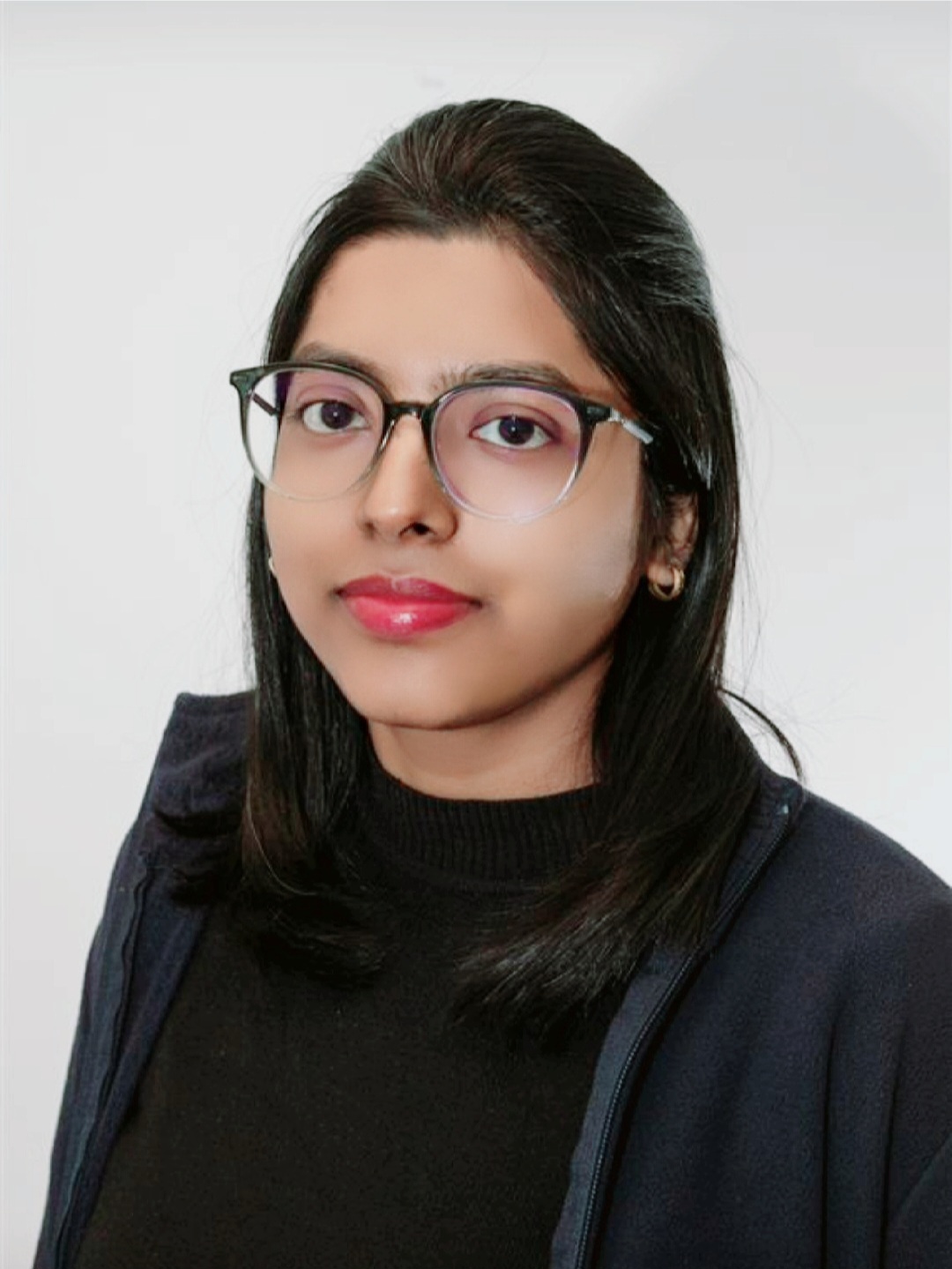}}]{Pratibha Verma} (Member, IEEE) is currently an Assistant Professor with the Department of Electrical Engineering, Indian Institute of Technology Indore, India. She received the Ph.D. degree in Electrical Engineering from Indian Institute of Technology Delhi, India. Prior to joining IIT Indore, she worked as a Postdoctoral Researcher with the Centre for Electronics Frontiers, University of Edinburgh, U.K. Her research interests include analog and mixed-signal integrated circuit design, energy-harvesting and power-management circuits, and AI-assisted electronic circuit design and automation.
    
\end{IEEEbiography}

\begin{IEEEbiography}
[{\includegraphics[width=1in,height=1.25in,clip,keepaspectratio]{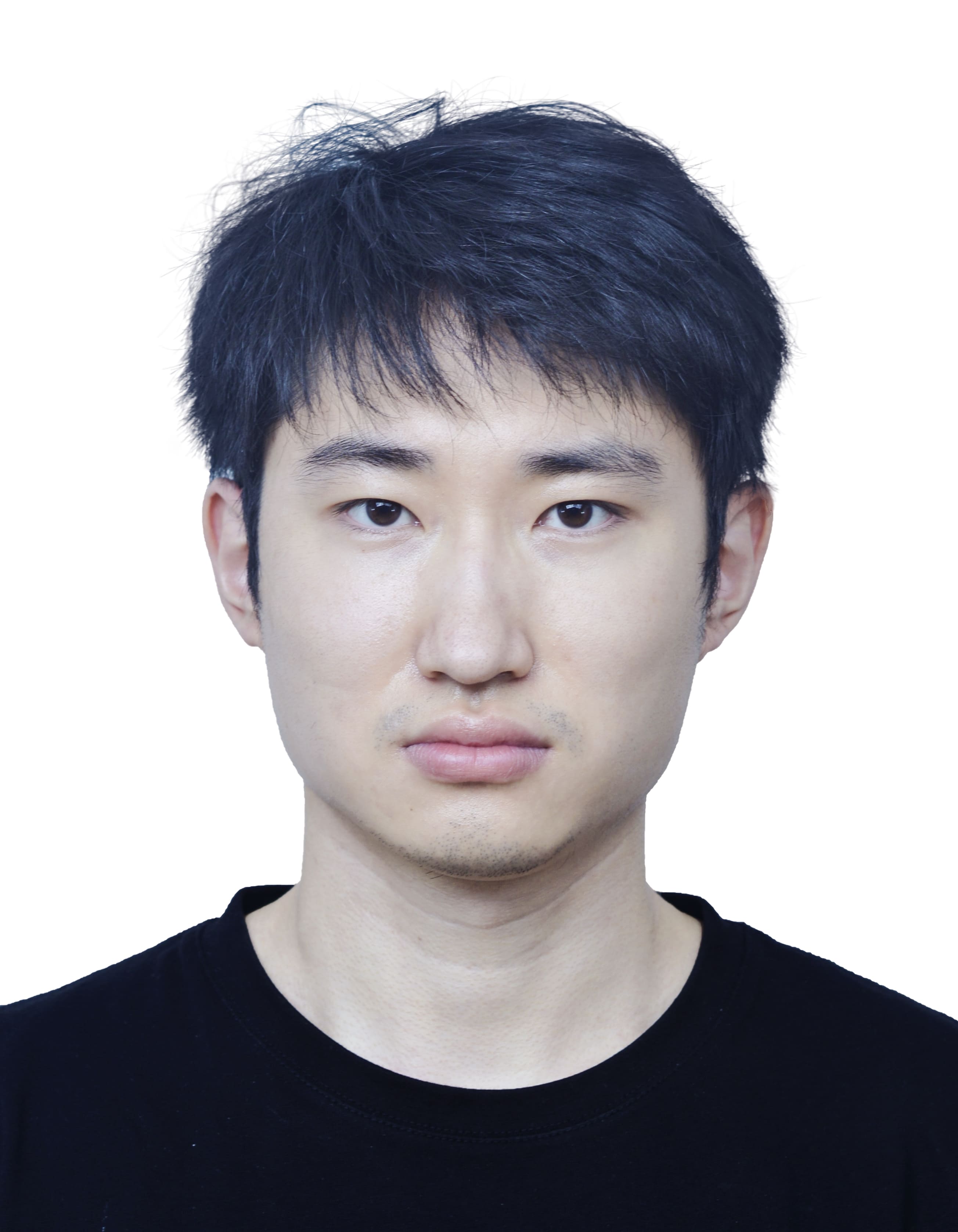}}]{Dongxu Guo} received the B.S. degree in electronics engineering from Beijing Normal University, Beijing, China, in 2021, and the M.Sc. degree in electronics from The University of Edinburgh, Edinburgh, U.K., in 2022. He is currently pursuing the Ph.D. degree with the School of Engineering, The University of Edinburgh, Edinburgh, U.K. His research interests include analog and mixed-signal integrated circuits, neuromorphic engineering, and memristor-based front-end circuits.
\end{IEEEbiography}

\begin{IEEEbiography}
[{\includegraphics[width=1in,height=1.25in,clip,keepaspectratio]{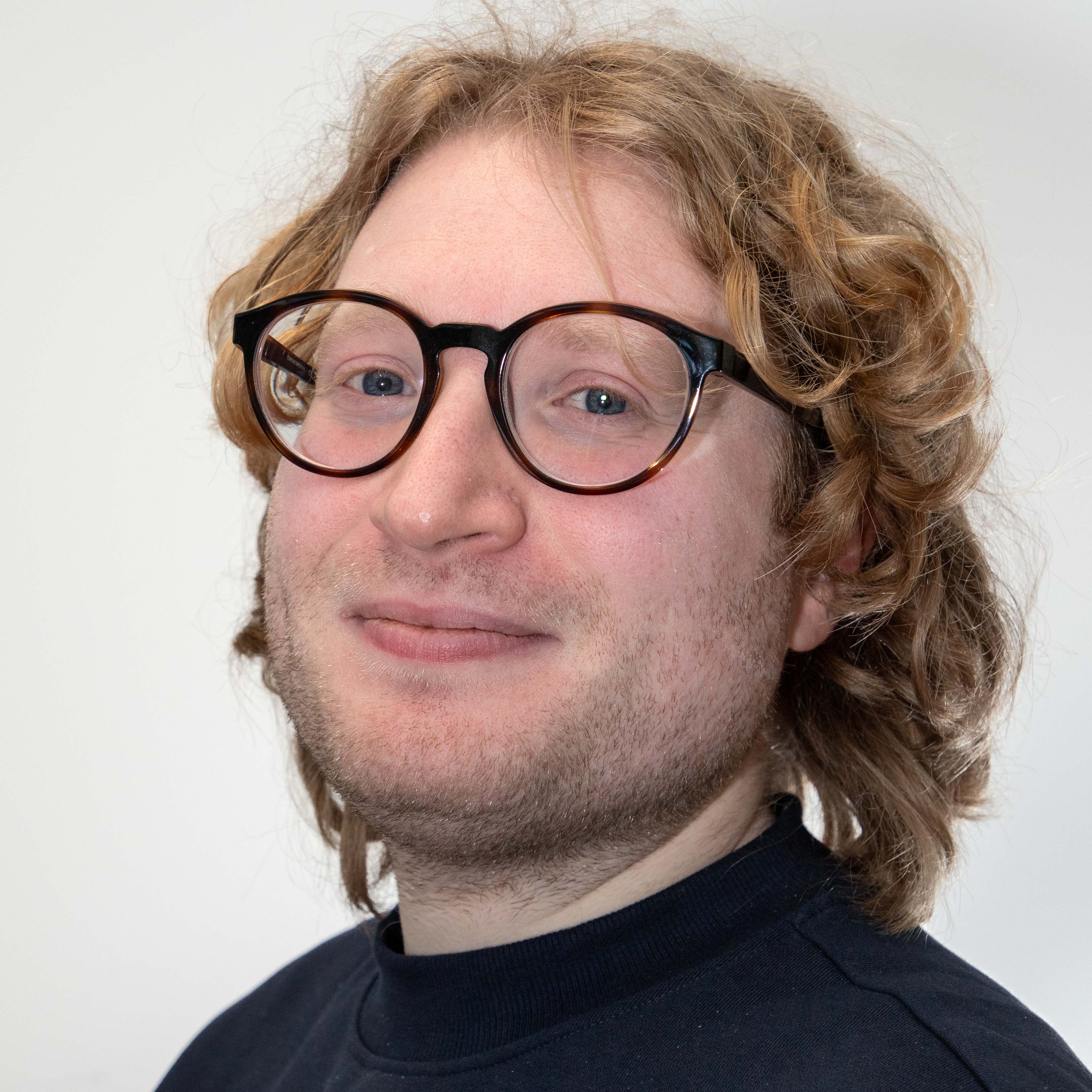}}]{Cristian Sestito}
(Member, IEEE) is a Research Fellow at the Centre for Electronics Frontiers, University of Edinburgh, U.K. He earned his B.Sc. (2016) and M.Sc. (2019) in Electronic Engineering, followed by a Ph.D. in Information and Communication Technologies (2023), from the University of Calabria, Italy. His PhD focused on the design and FPGA implementation of convolutional neural networks, with particular emphasis on dilated and transposed convolutions. During 2021–2022, he was a Visiting Scholar at Heriot-Watt University, Edinburgh, U.K. where he worked on neural network compression techniques. His current research interests include digital AI accelerator design, software simulators for neuromorphic AI, and AI-assisted electronic design automation frameworks.
\end{IEEEbiography}

\begin{IEEEbiography}
[{\includegraphics[width=1in,height=1.25in,clip,keepaspectratio]{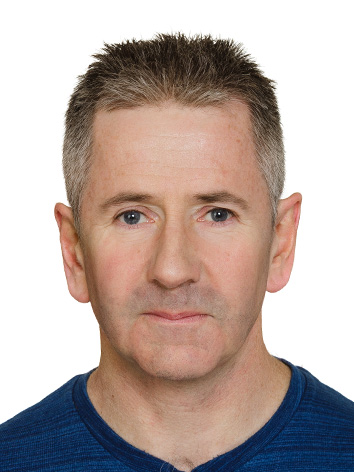}}]{Michael O'Boyle}
(Fellow, IEEE) FBCS FRSE is a Professor of Computer Science at the University of Edinburgh. He is best known for his work in automatic program optimisation, incorporating machine learning into compilation and parallelisation. Over his career, he has been awarded three fellowships, five international visiting positions and eleven best paper awards. His current work focuses on matching  accelerator hardware to legacy code using program synthesis and neural machine translation
\end{IEEEbiography}

\begin{IEEEbiography}
[{\includegraphics[width=1in,height=1.25in,clip,keepaspectratio]{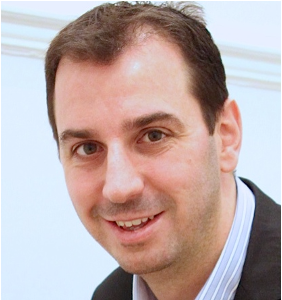}}]{Christos-Savvas Bouganis}
(Senior Member, IEEE) is a Professor of Intelligent Digital Systems in the Department of Electrical and Electronic Engineering, Imperial College London, U.K. He is leading the iDSL group at Imperial College (https://www.imperial.ac.uk/idsl), with a focus on the theory and practice of reconfigurable computing and design automation, mainly targeting the domains of Machine Learning, Computer Vision, and Robotics.
\end{IEEEbiography}

\begin{IEEEbiography}
[{\includegraphics[width=1in,height=1.25in,clip,keepaspectratio]{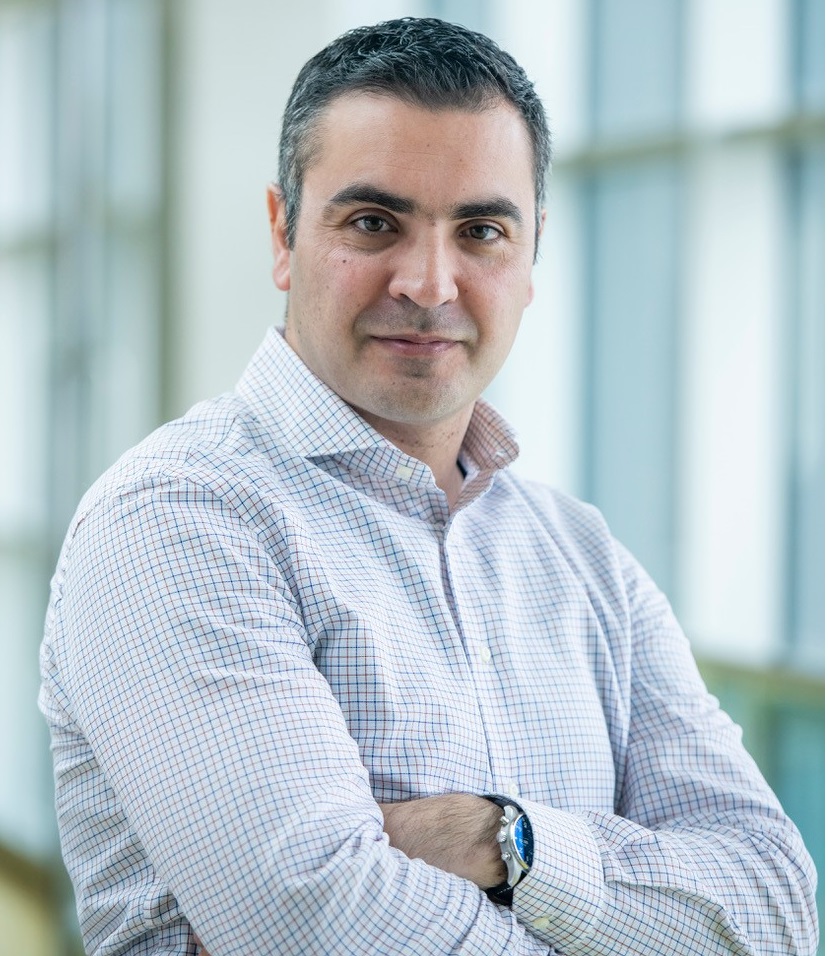}}]{Themis Prodromakis}
(Senior Member, IEEE) received the bachelor’s degree in electrical and electronic engineering from the University of Lincoln, U.K., the M.Sc. degree in microelectronics and telecommunications from the University of Liverpool, U.K., and the Ph.D. degree in electrical and electronic engineering from Imperial College London, U.K. He then held a Corrigan Fellowship in nanoscale technology and science with the Centre for Bio-Inspired Technology, Imperial College London, and a Lindemann Trust Visiting Fellowship with the Department of Electrical Engineering and Computer Sciences, University of California at Berkeley, USA. He was a Professor of nanotechnology at the University of Southampton, U.K. He holds the Regius Chair of Engineering at the University of Edinburgh. He is currently a Royal Academy of Engineering Chair in emerging technologies and a Royal Society Industry Fellowship. His background is in electron devices and nanofabrication techniques. His current research interests include memristive technologies for advanced computing architectures and biomedical applications. He is a fellow of the Royal Society of Chemistry, the British Computer Society, the IET, and the Institute of Physics.
\end{IEEEbiography}

\vfill

\end{document}